\documentclass[journal]{IEEEtran}
\usepackage{amsmath,amsfonts}
\usepackage{url}
\usepackage{marvosym}
\usepackage{balance}
\usepackage[hidelinks,hypertexnames=false,colorlinks=true,linkcolor=blue!80!black,citecolor=red!70!black,urlcolor=black]{hyperref}
\usepackage{graphicx}
\usepackage{cite}
\usepackage{booktabs}
\usepackage{multirow}
\usepackage{threeparttable}
\usepackage{makecell}
\usepackage{mathrsfs}
\usepackage{enumitem}
\usepackage{ulem}
\usepackage{color}
\usepackage{float}
\usepackage{caption}

\usepackage[switch]{lineno}

\newcommand{\settablefont}{\fontsize{6}{12}\selectfont}

\usepackage[table,xcdraw]{xcolor}
\usepackage{scalerel}
\usepackage{tikz}
\usetikzlibrary{svg.path}

\definecolor{orcidlogocol}{HTML}{A6CE39}
\tikzset{
  orcidlogo/.pic={
    \fill[orcidlogocol] svg{M256,128c0,70.7-57.3,128-128,128C57.3,256,0,198.7,0,128C0,57.3,57.3,0,128,0C198.7,0,256,57.3,256,128z};
    \fill[white] svg{M86.3,186.2H70.9V79.1h15.4v48.4V186.2z}
                 svg{M108.9,79.1h41.6c39.6,0,57,28.3,57,53.6c0,27.5-21.5,53.6-56.8,53.6h-41.8V79.1z M124.3,172.4h24.5c34.9,0,42.9-26.5,42.9-39.7c0-21.5-13.7-39.7-43.7-39.7h-23.7V172.4z}
                 svg{M88.7,56.8c0,5.5-4.5,10.1-10.1,10.1c-5.6,0-10.1-4.6-10.1-10.1c0-5.6,4.5-10.1,10.1-10.1C84.2,46.7,88.7,51.3,88.7,56.8z};
  }
}
\newcommand\orcidicon[1]{\href{https://orcid.org/#1}{\mbox{\scalerel*{
\begin{tikzpicture}[yscale=-1,transform shape]
\pic{orcidlogo};
\end{tikzpicture}
}{|}}}}
\newcolumntype{L}[1]{>{\raggedright\arraybackslash}p{#1}}
\newcolumntype{C}[1]{>{\centering\arraybackslash}p{#1}}
\newcolumntype{R}[1]{>{\raggedleft\arraybackslash}p{#1}}

\definecolor{TYS}{rgb}{0.6, 0.8, 0.2}
\definecolor{DOcolor}{rgb}{1,0.45,0.0}
\definecolor{NAVYcolor}{rgb}{0.05,0,0.5}

\begin{document}
\title{DCPI-Depth: Explicitly Infusing Dense Correspondence Prior to Unsupervised \\ Monocular Depth Estimation}

\normalem
\author{Mengtan Zhang$^{\orcidicon{0009-0003-3468-7680}\,}$, Yi Feng$^{\orcidicon{0009-0005-4885-0850}\,}$, Qijun Chen$^{\orcidicon{0000-0001-5644-1188}\,}$,~\IEEEmembership{Senior Member,~IEEE}, and Rui Fan$^{\orcidicon{0000-0003-2593-6596}\,}$,~\IEEEmembership{Senior Member,~IEEE}
\thanks{
This research was supported by the Science and Technology Commission of Shanghai Municipal under Grant 22511104500, the National Natural Science Foundation of China under Grant 62233013, the Fundamental Research Funds for the Central Universities, and Xiaomi Young Talents Program. (\emph{Corresponding author: Rui Fan})}
\thanks{Mengtan Zhang, Yi Feng, Qijun Chen, and Rui Fan are with the College of Electronics \& Information Engineering, Shanghai Research Institute for Intelligent Autonomous Systems, Shanghai Institute of Intelligent Science and Technology, the State Key Laboratory of Intelligent Autonomous Systems, and Frontiers Science Center for Intelligent Autonomous Systems, Tongji University, Shanghai 201804, China (e-mails: 2050026@tongji.edu.cn, fengyi@ieee.org, \{qjchen, rfan\}@tongji.edu.cn).}
}

\maketitle

\begin{abstract}
There has been a recent surge of interest in learning to perceive depth from monocular videos in an unsupervised fashion. 
A key challenge in this field is achieving robust and accurate depth estimation in challenging scenarios, particularly in regions with weak textures or where dynamic objects are present. This study makes three major contributions by delving deeply into dense correspondence priors to provide existing frameworks with explicit geometric constraints. The first novelty is a contextual-geometric depth consistency loss, which employs depth maps triangulated from dense correspondences based on estimated ego-motion to guide the learning of depth perception from contextual information, since explicitly triangulated depth maps capture accurate relative distances among pixels. The second novelty arises from the observation that there exists an explicit, deducible relationship between optical flow divergence and depth gradient. A differential property correlation loss is, therefore, designed to refine depth estimation with a specific emphasis on local variations. The third novelty is a bidirectional stream co-adjustment strategy that enhances the interaction between rigid and optical flows, encouraging the former towards more accurate correspondence and making the latter more adaptable across various scenarios under the static scene hypotheses. DCPI-Depth, a framework that incorporates all these innovative components and couples two bidirectional and collaborative streams, achieves state-of-the-art performance and generalizability across multiple public datasets, outperforming all existing prior arts. Specifically, it demonstrates accurate depth estimation in texture-less and dynamic regions, and shows more reasonable smoothness. Our source code will be publicly available at mias.group/DCPI-Depth upon publication. 
\end{abstract}


\section{Introduction}
\label{Sect.intro}

\IEEEPARstart{M}{onocular} depth estimation, a crucial research field in computer vision and robotics, has applications across various domains, such as autonomous driving \cite{geiger2013vision}, augmented reality \cite{luo2020consistent}, and embodied artificial intelligence \cite{wang2024visual}.
It provides agents with powerful environmental perception capabilities, enabling robust ego-motion estimation and 3D geometry reconstruction \cite{zhou2017unsupervised}.
Early monocular depth estimation approaches \cite{eigen2014depth, liu2015learning}, developed based on supervised learning, typically require a large amount of well-annotated, per-pixel depth ground truth (generally acquired using high-precision LiDARs) for model training \cite{zhou2021self, zhao2022monovit}. 
Nevertheless, collecting and labeling such data is tedious and costly \cite{godard2019digging, zhou2021self}. Thus, the practical use of these supervised approaches remains limited.

In recent years,  un/self-supervised monocular depth estimation has garnered significant attention \cite{zhou2017unsupervised, godard2019digging, xu2021multi, sun2023sc, zhang2023lite}. Such approaches obviate the need for extensive depth ground truth by leveraging either stereo image pairs 
\cite{godard2017unsupervised, ye2021unsupervised} or monocular videos to jointly learn depth and ego-motion estimation \cite{zhou2017unsupervised}. 
Specifically, given the target and source images, the estimated depth map (at the target view) and camera ego-motion are employed to warp the source image into the target view. The depth estimation network (hereafter referred to as \textit{DepthNet}) and the pose estimation network (hereafter referred to as \textit{PoseNet}) are then jointly trained in an un/self-supervised manner by minimizing the photometric loss, which measures the consistency between the reconstructed and original target images \cite{zhou2017unsupervised, godard2019digging, sun2023sc}. 

Despite the progress made, three limitations in existing frameworks continue to impede further advances in monocular depth estimation:
\begin{enumerate}[label=(\arabic*)]
    \item DepthNet perceives depth based on the contextual information in RGB images. While it effectively determines whether an object is in front of or behind another, accurately and efficiently learning their relative distance by minimizing the photometric loss is challenging. This is because photometric loss cannot directly reflect the magnitude of depth error, sometimes resulting in unsuitable gradients for optimizing depth during backpropagation.
    \item Local depth variation is commonly constrained by edge-aware smoothness loss, which encourages local smoothness in depth based on image gradients \cite{godard2017unsupervised}. However, since changes in image intensity do not directly correlate with local depth variation, this enforced smoothing can introduce errors in depth estimation.
    \item 
    Comparable photometric losses can result from different rigid flows, which may map a pixel to the wrong candidates with similar pixel intensities. This implies that the supervisory signals provided by the photometric loss to DepthNet and PoseNet are indirect, possibly leading to unsatisfactory robustness of monocular depth estimation, particularly in regions with weak textures.
\end{enumerate}

Therefore, in this article, we introduce a novel unsupervised monocular depth estimation framework, known as \uline{\textbf{D}ense \textbf{C}orrespondence \textbf{P}rior-\textbf{I}nfused \textbf{Depth} \textbf{(DCPI-Depth)}}, to overcome the limitations above by exploiting the depth cues in dense correspondence priors.
Our DCPI-Depth consists of two bidirectional and collaborative streams: a traditional photometric consistency-guided (PCG) stream and our proposed correspondence prior-guided (CPG) stream. The PCG stream, following the prevalently used methods \cite{zhou2017unsupervised, godard2019digging, sun2023sc, zhang2023lite}, employs estimated depth and ego-motion information to warp the source frame into the target view, and then computes a photometric loss to provide the supervisory signal. On the other hand, the CPG stream leverages a pre-trained FlowNet \cite{teed2020raft} to provide dense correspondence priors. These priors are first utilized along with the estimated ego-motion to construct a geometric-based depth map via triangulation \cite{hartley1997triangulation}. Such an explicitly derived depth map captures accurate relative distances among pixels. By enforcing consistency between these two sources of depth maps through a newly developed contextual-geometric depth consistency (CGDC) loss, we significantly optimize the convergence of DepthNet during training. Moreover, optical flow divergence, a differential property of dense correspondence priors, is found to have an explicit relationship with depth gradient. Building upon this relationship, we develop a novel differential property correlation (DPC) loss to improve depth quality from the aspect of local variation. Finally, a bidirectional stream co-adjustment (BSCA) strategy is adopted to make the two streams complement each other, where the rigid flow in the PCG stream mainly alleviates the misguidance of the CPG steam to depth on dynamic objects, while the optical flow in the CPG stream refines the rigid flow of the PCG stream with dense correspondences.

In summary, the main contributions of this article include:
\begin{itemize}
    \item DCPI-Depth, a novel unsupervised monocular depth estimation framework with a CPG stream developed to infuse dense correspondence prior into the traditional PCG stream; 
    \item A CGDC loss to optimize the convergence of DepthNet using a geometric-based depth map constructed by triangulating dense correspondence priors with estimated ego-motion;
    \item A DPC loss to further refine the quality of the estimated depth from the aspect of local variation based on the explicit relationship between optical flow divergence and depth gradient;
    \item A BSCA strategy to enable the two streams to complement each other, with a specific emphasis on improving depth accuracy in dynamic regions without masking technique.
    \end{itemize}

The remainder of this article is organized as follows:
Sect. \ref{Sect.related_work} presents an overview of the existing monocular depth estimation methods. In Sect. \ref{Sect.methodology}, we detail the proposed DCPI-Depth framework. In Sect. \ref{Sect.experiments}, we present the experimental results across several public datasets. 
Finally, we conclude this article and discuss possible future work in Sect. \ref{Sect.conclusion}.

\section{Related Work}
\label{Sect.related_work}
\subsection{Supervised Monocular Depth Estimation}
\label{sec.supervised_related}

Supervised monocular depth estimation approaches require depth ground truth for model training. As the first attempt, the study \cite{eigen2014depth} proposes a coarse-to-fine architecture and a scale-invariant loss function to perceive depth from a single image. Subsequent research focuses mainly on improving depth estimation performance by exploring more intricate network architectures \cite{cao2017estimating, liu2015learning} or developing new loss functions \cite{hu2019revisiting, fu2018deep}. For example, in studies \cite{cao2017estimating} and \cite{fu2018deep}, monocular depth estimation was reformulated as a per-pixel classification task, where depth ranges instead of exact depth values are predicted. Furthermore, to combine the benefits of both regression and classification tasks, the study \cite{bhat2021adabins} redefines this problem as a per-pixel classification-regression task. In \cite{lee2019big}, multi-scale guidance layers are introduced to establish connections between intermediate-layer features and the final depth map. In \cite{yang2021transformer}, a vision Transformer (ViT)-based architecture was developed to capture long-range correlations in depth estimation. More recently, Depth Anything \cite{yang2024depth} has demonstrated impressive performance, primarily due to its powerful backbone (a ViT-based vision foundation model) that is capable of extracting general-purpose, informative deep features. It first reproduces a MiDaS-based \cite{birkl2023midas} teacher model with pre-trained weights from DINOv2 \cite{oquab2024dinov}, and then utilizes the teacher's predictions as pseudo-labels to train a student model on large-scale unlabeled data. Building on single-image depth estimation, monocular video depth estimation incorporates both temporal and geometric consistencies \cite{luo2020consistent}. Representative works, such as \cite{luo2020consistent} and \cite{kopf2021robust}, pioneer the use of correspondences and camera poses to enforce inter-frame depth consistency in 3D space, which inspires us to further explore the way of infusing dense correspondence prior to guide the unsupervised learning for monocular depth estimation.

\subsection{Un/Self-Supervised Monocular Depth Estimation}
\label{sec.unsupervised_related}
To liberate monocular depth estimation from dependence on extensive ground-truth data, un/self-supervised approaches \cite{godard2017unsupervised, zhou2017unsupervised, godard2019digging, sun2023sc, zhang2023lite} have emerged as visible alternatives. These methods typically utilize estimated depth to establish a differentiable warping between two images and employ photometric loss as the supervisory signal \cite{garg2016unsupervised, zhou2017unsupervised}. The study \cite{garg2016unsupervised} represents the first reported attempt to learn monocular depth estimation from stereo image pairs within a self-supervised framework. Subsequently, the study \cite{zhou2017unsupervised} extends this approach by coupling the learning of depth and ego-motion estimation from monocular videos.
However, un/self-supervised methods often face challenges with independently moving objects and preserving clear object boundaries, due to multi-view ambiguities \cite{sun2023sc}. Therefore, in \cite{godard2019digging}, a minimum reprojection loss and an auto-masking technique were introduced to exclude such regions during model training, significantly improving depth estimation performance. Building upon these prior arts, several studies have explored more sophisticated network architectures \cite{zhang2020unsupervised, guizilini20203d, zhang2022self, zhang2023lite, han2023self} for improved performance. Others incorporate additional relevant tasks such as optical flow estimation \cite{zou2018df, yin2018geonet, ranjan2019competitive} and semantic segmentation \cite{chen2023self, jung2021fine} to enhance cross-task consistency or address the dynamic object challenge.

While these efforts have demonstrated promising performance, the existing unsupervised frameworks still present significant opportunities for refinement \cite{sun2023sc}. This limitation primarily arises from the reliance on contextual information to infer the pixel-wise depth map, which is indirectly supervised through the minimization of photometric loss. In the absence of guidance from additional and meaningful prior knowledge, DepthNet struggles to perform robustly in challenging scenarios. Therefore, in study \cite{sun2023sc}, a monocular depth estimation model pre-trained on large-scale datasets is utilized to provide pseudo depth, a single-image depth prior, and two depth refinement loss functions are also designed to achieve more robust and reliable depth estimation. However, the limited accuracy of the pseudo depth significantly restricts the refinement capabilities of these loss functions. Therefore, in this article, we resort to dense correspondence priors for depth refinement. Unlike pseudo depth, these priors offer more direct, reliable, and interpretable geometric guidance through our developed CGDC and DPC losses.

\section{Methodology}
\label{Sect.methodology}

\subsection{Overall Architecture}
\label{Sect.architecture}

\begin{figure*}[t!]
	\centering
	\includegraphics[width=0.99 \textwidth]{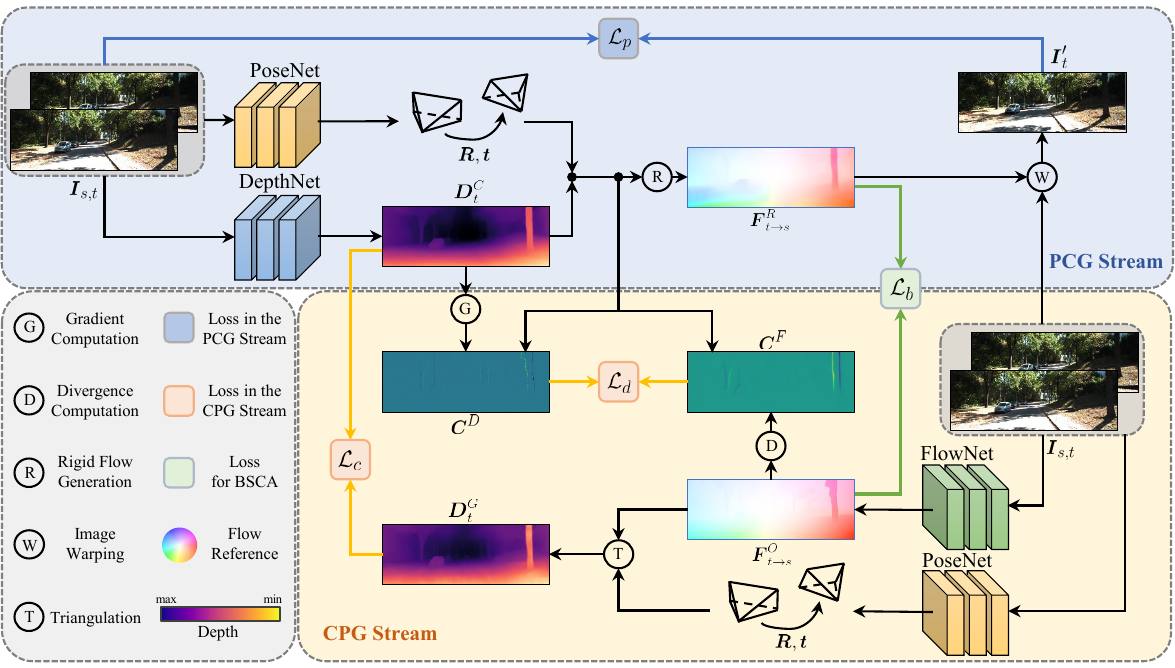}
	\caption{The overall architecture of our proposed DCPI-Depth framework, which consists of two collaborative and bidirectional streams: PCG and CPG. The input image pairs, PoseNet, and the estimated ego-motion are depicted separately in each stream.
    }
\label{fig.overall_architect}
\end{figure*}

As illustrated in Fig. \ref{fig.overall_architect}, our proposed DCPI-Depth framework comprises two collaborative and bidirectional streams: PCG and CPG. The former, following the prior studies \cite{godard2019digging, sun2023sc, zhang2023lite}, effectively yet indirectly supervises the training of DepthNet and PoseNet through photometric loss, while the latter infuses dense correspondence priors (provided by a pre-trained FlowNet) into the former to overcome the limitations of current SoTA frameworks \cite{sun2023sc, zhang2023lite, han2023self}, which rely solely on the PCG stream. Specifically, within the CPG stream, we introduce two novel loss functions: 
(1) a CGDC loss that guides the training of DepthNet by enforcing consistency between geometric-based and contextual-based depth maps, enabling DepthNet to capture accurate relative distances among pixels, thereby optimizing its convergence during training;
(2) a DPC loss to constrain the local variation of depth based on the explicit relationship between optical flow divergence and depth gradient.
Furthermore, we develop a BSCA strategy to collectively improve the aforementioned two streams: the rigid flow ensures accurate geometric guidance for depth estimation mainly on dynamic objects within the CPG stream, while the optical flow refines the rigid flow with dense correspondence within the PCG stream, thus allowing these two streams to effectively complement each other.

\subsection{Contextual-Geometric Depth Consistency Loss}
\label{Sect.DepthConsistency}

In the conventional PCG stream, given target and source video frames\footnote{In this article, the subscripts ``$t$" and ``$s$" denote ``target" and ``source", respectively.} $\boldsymbol{I}_{t, s} \in \mathbb{R}^{H \times W \times 3}$, DepthNet takes $\boldsymbol{I}_{t}$ as input to infer a depth map $\boldsymbol{D}^C_t \in \mathbb{R}^{H \times W}$ based on contextual information, where $H$ and $W$ represent the height and width of the input image, respectively, while PoseNet estimates the ego-motion, including a rotation matrix $\boldsymbol{R}=[\boldsymbol{r}_1^{\top}, \boldsymbol{r}_2^{\top}, \boldsymbol{r}_3^{\top}]^\top \in SO(3)$ and 
a translation vector $\boldsymbol{t}=[t_1,t_2,t_3]^\top \in \mathbb{R}^{3}$ from the source view to the target view. 
The rigid flow map $\boldsymbol{F}_{t \to s}^R \in \mathbb{R}^{H \times W \times 2}$ from the target view to the source view can then be generated as follows:
\begin{equation}
    \begin{bmatrix}\boldsymbol{F}^R_{t\to s}(\boldsymbol{p}_t)  \\ 0 \end{bmatrix} + \tilde{\boldsymbol{p}}_t  \sim \boldsymbol{K} \begin{bmatrix}\boldsymbol{R} & \boldsymbol{t} \end{bmatrix}\begin{bmatrix}\boldsymbol{D}^C_t(\boldsymbol{p}_t)\boldsymbol{K}^{-1}\tilde{\boldsymbol{p}}_t \\ 1\end{bmatrix},
    \label{eq:fr2}
\end{equation}
where the symbol $\sim $ indicates that two vectors are equal up to a scale factor,  
$\boldsymbol{K}$ represents the camera intrinsic matrix, 
$\boldsymbol{p}_t=[u,v]^\top$ denotes a 2D pixel, and $\tilde{\boldsymbol{p}}_t$ is its homogeneous coordinates. $\boldsymbol{I}_s$ is then warped into the target view using the rigid flow map $\boldsymbol{F}_{t \to s}^R$, generating $\boldsymbol{I}' _t$. By comparing $\boldsymbol{I}' _t$ with $\boldsymbol{I}_t$,
the following photometric loss is computed to provide a supervisory signal for the training of both DepthNet and PoseNet \cite{bian2021unsupervised}:
\begin{equation}
    \mathcal{L}_p =  \alpha \frac{1-SSIM(\boldsymbol{I}' _t,\boldsymbol{I}_t)}{2} +(1-\alpha )\left \| \boldsymbol{I}' _t-\boldsymbol{I}_t \right \|_1 ,
    \label{eq.photometric_loss}
\end{equation}
where $SSIM$ denotes the pixel-wise structural similarity index operation \cite{wang2004image}, and $\alpha$ is an empirical weight set to 0.85.

DepthNet in the conventional PCG stream is trained to infer depth value per pixel from contextual information by minimizing (\ref{eq.photometric_loss}) based on given RGB image pairs. Previous studies have neglected to incorporate geometric guidance into DepthNet training, leading to a significant limitation. While DepthNet can ascertain whether an object is in front of or behind another relative to the camera origin from contextual information, it struggles to effectively learn the extent of their relative distances solely by minimizing the photometric loss. This challenge arises because errors in image intensities do not directly reflect depth errors in terms of magnitude, rendering the gradient from photometric loss during back-propagation not always suitable for depth optimization.

To address these limitations, we resort to dense correspondence priors to generate another depth map based on well-developed and interpretable principles of multi-view geometry, thereby providing an additional constraint on depth estimation from RGB images. 
We first introduce a pre-trained FlowNet \cite{teed2020raft} to generate the optical flow map $\boldsymbol{F}_{t \to s}^O \in \mathbb{R}^{H \times W \times 2}$, from which the following dense correspondence priors are derived:
\begin{align}
    \begin{cases}
    \hat{\boldsymbol{p}}_t^C = \boldsymbol{K}^{-1} \tilde{\boldsymbol{p}}_t \\[0.6ex]
    \hat{\boldsymbol{p}}_s^C = \boldsymbol{K}^{-1}\left(\tilde{\boldsymbol{p}}_t+\begin{bmatrix}\boldsymbol{F}_{t\to s}^O(\boldsymbol{p}_t) \\ 0 \end{bmatrix}\right), 
\end{cases}
\end{align}
where $\hat{\boldsymbol{p}}_t^C$ and $\hat{\boldsymbol{p}}_s^C$ represent a pair of normalized camera coordinates along the optical axis in the target and the source views, respectively. We then leverage such dense correspondence priors along with the ego-motion estimated by PoseNet to construct a geometric-based depth map $\boldsymbol{D}^G_t \in \mathbb{R}^{H \times W}$ via triangulation \cite{hartley1997triangulation} based on the following relationship:
\begin{equation}
    \hat{\boldsymbol{p}}_s^C \sim  \begin{bmatrix}\boldsymbol{R} & \boldsymbol{t} \end{bmatrix} \begin{bmatrix}\boldsymbol{D}^G _t(\boldsymbol{p}_t)\hat{\boldsymbol{p}}_t^C \\ 1\end{bmatrix}.
    \label{eq:hat_p_1}
\end{equation}
$\hat{p}_{s, i}^C$, the $i$-th element in $\hat{\boldsymbol{p}}_s^C$ ($i=\{1, 2\}$), is expressed as follows:
\begin{equation}
\label{eq:hat_p_2}
    \hat{p}_{s, i}^C=\frac{\boldsymbol{D}^G _t(\boldsymbol{p}_t) \boldsymbol{r}_i^{\top}\hat{\boldsymbol{p}}_t^C + t_i}{\boldsymbol{D}^G _t(\boldsymbol{p}_t) \boldsymbol{r}_3^{\top}\hat{\boldsymbol{p}}_t^C + t_3}.
\end{equation}
$\boldsymbol{D}^G _t$ can then be yielded as follows:
\begin{equation}
\label{eq:d^trf}
    \boldsymbol{D}^G _t(\boldsymbol{p}_t) = \frac{ \displaystyle\sum_{i=1}^{2}\left (   t_i - \hat{p}_{s, i}^C t_3\right )  }{ \displaystyle\sum_{i=1}^{2}\left ( \hat{p}_{s, i}^C \boldsymbol{r}_3^{\top}\hat{\boldsymbol{p}}_t^C-\boldsymbol{r}_i^{\top}\hat{\boldsymbol{p}}_t^C \right )  }.  
\end{equation}
Finally, we employ the following CGDC loss:
\begin{equation}
\label{eq.lcgdc}
    \mathcal{L}_{c} = \frac{1}{HW}  \sum_{ \boldsymbol{p}  } \frac{\left | \boldsymbol{D}^G _t(\boldsymbol{p}) - \boldsymbol{D}^C_t(\boldsymbol{p}) \right |} {\boldsymbol{D}^C_t(\boldsymbol{p}) }
\end{equation}
to provide DepthNet with an additional constraint, enabling it to capture accurate relative distances among pixels. Considering that higher depth values potentially exhibit greater absolute depth error, we adopt relative error in our CGDC loss to ensure more consistent gradients for back-propagation across all pixels. The effectiveness of our proposed CGDC loss is validated and discussed in Sect. \ref{section:abl}.

\begin{figure*}[t!]
	\centering
	\includegraphics[width=0.99\textwidth]{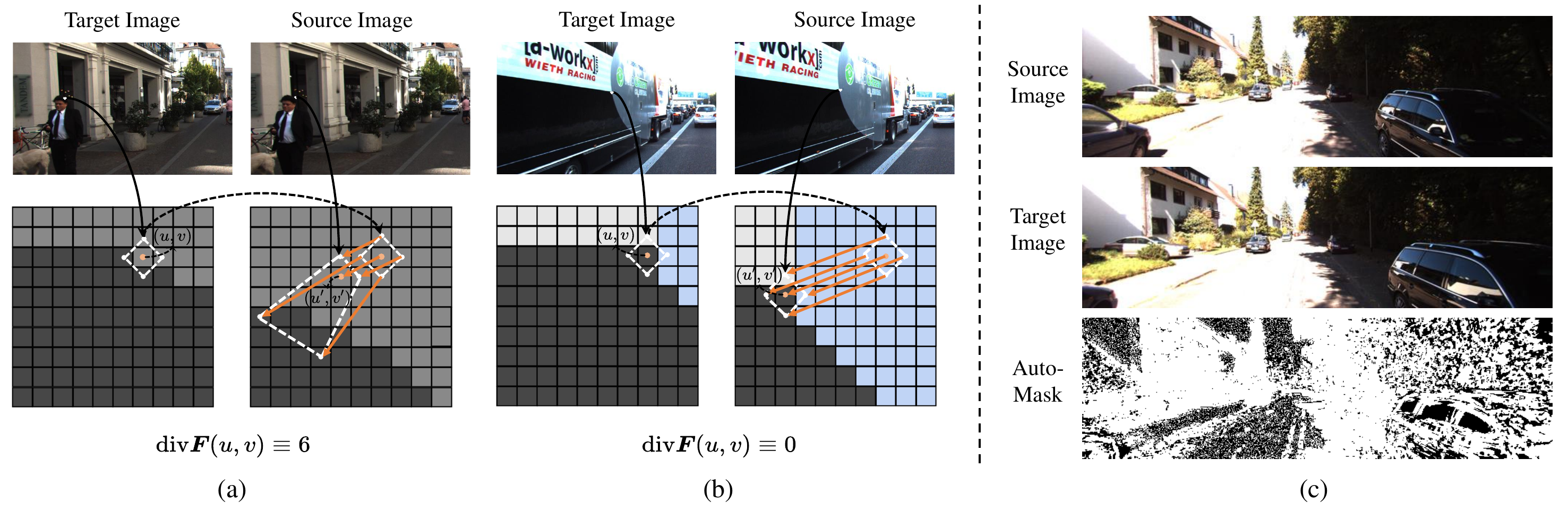}
	\caption{Illustrations of optical flow divergence and auto-masking result: (a) optical flow divergence for pixels with similar intensities yet being spatially discontinuous; (b) optical flow divergence for pixels with significantly different intensities yet being spatially continuous; (c) auto-masking result for the given source and target images. A given pixel and its four neighbors in the target image are utilized for visualization in (a) and (b), where it can be observed that their correspondences in the source image are widely separated in (a) but similarly distributed in (b). The auto-masking algorithm tends to overly mask static regions, particularly in low-texture areas or when overexposed, and cannot effectively mask dynamic objects.}
\label{fig.motivation}
\end{figure*}

\subsection{Differential Property Correlation Loss}
\label{Sect.DifferentialAlignment}
Existing approaches \cite{zhang2023lite, han2023self} that include only the PCG stream often struggle to distinguish and effectively handle regions with different levels of continuity. Specifically, these methods encounter difficulties in ensuring smooth depth changes in continuous regions and preserving clear boundaries near or at discontinuities. This problem arises primarily due to the lack of proper constraints that encourage DepthNet to consider local depth variations, especially since the depth of each pixel is estimated independently. Several studies \cite{godard2019digging, bian2021unsupervised, lyu2021hr, zhou2021self, zhang2023lite, han2023self} introduced an edge-aware smoothness loss based on image gradients, initially presented in \cite{godard2017unsupervised}, to encourage local smoothness in depth estimation. However, such a loss function is somewhat problematic and incomplete. While this loss function is effective in regions where depth and image intensity have consistent change trends, it cannot constrain the extent of smoothing. Moreover, in continuous regions with rich texture or at discontinuities with subtle texture changes, this loss function may prove ineffective or even cause misleading guidance.  

Compared to pixel intensity, it has been found that the dense correspondence priors between two images, as provided by FlowNet, have a more direct relationship with depth changes. As illustrated in Fig. \ref{fig.motivation}(a), 
adjacent pixels that have similar intensities but are spatially discontinuous (because they are located on different objects) are likely to exhibit significantly different apparent motions, resulting in high optical flow divergence due to the separation of these pixels. In contrast, as illustrated in Fig. \ref{fig.motivation}(b), pixels that have different intensities but are located in continuous regions typically have similar apparent motions, resulting in low optical flow divergence. ${\rm div} \boldsymbol{F}$, the divergence of the given optical flow $\boldsymbol{F}$ can be numerically calculated through the following expression: 
\begin{align}  
{\rm div} \boldsymbol{F}(u, v) \equiv & -\boldsymbol{n}_u^\top \boldsymbol{F}(u-1,v) + \boldsymbol{n}_u^\top \boldsymbol{F}(u+1,v) \\[0.8ex] \notag
&+ \boldsymbol{n}_v^\top \boldsymbol{F}(u,v+1) - \boldsymbol{n}_v^\top \boldsymbol{F}(u,v-1),
\label{eq.div}
\end{align}
where $\boldsymbol{n}_u=[1, 0]^\top$ and $\boldsymbol{n}_v=[0, 1]^\top$ are two unit vectors in the horizontal and vertical directions, and $\equiv $ represents discretization. Therefore, we establish an explicit constraint on local depth variation by leveraging the two correlated differential properties: optical flow divergence and depth gradient. This allows for more accurate depth estimation by ensuring that changes in depth are consistently aligned with variations in optical flow.
It has been proven in the study \cite{bian2021auto} that rotational flow is independent of depth. Therefore, only the translational component of ego-motion contributes to constraining depth estimation from the aspect of local variation using optical flow divergence, and incorporating rotational flow, particularly when it is substantial, can disrupt this constraint. To address this issue, we eliminate the rotational apparent motions in the optical flow using the following expression:
\begin{equation}
    \boldsymbol{F}^{\rm Tra}_{t\to s} = \boldsymbol{F}^O_{t\to s} - {\boldsymbol{F}^{\rm{Rot}}_{t\to s}}, 
\end{equation}
where $\boldsymbol{F}^{\rm Tra}_{t\to s}$ denotes the translational optical flow, and $\boldsymbol{F}^{\rm Rot}_{t\to s}$, the rotational rigid flow, is generated using the estimated $\boldsymbol{R}$ and a translation vector of zeros $\boldsymbol{0}$ as follows:
\begin{equation}
    \begin{bmatrix} \boldsymbol{F}^{\rm Rot}_{t\to s}(\boldsymbol{p}_t)  \\ 0 \end{bmatrix} + \tilde{\boldsymbol{p}}_t  \sim \boldsymbol{K} \begin{bmatrix}\boldsymbol{R} & \boldsymbol{0} \end{bmatrix}\begin{bmatrix}\boldsymbol{D}^C_t(\boldsymbol{p}_t)\boldsymbol{K}^{-1}\tilde{\boldsymbol{p}}_t \\ 1\end{bmatrix}.
\end{equation}
The relation between translational apparent motion and depth can then be written as follows:
\begin{equation}
\label{eq:appro_f}
    \begin{aligned}
     \begin{bmatrix} \boldsymbol{F}^{\rm Tra}_{t\to s}(\boldsymbol{p}_t) \\ 0\end{bmatrix} 
     & = \displaystyle\frac{\boldsymbol{K} \begin{bmatrix}\boldsymbol{I} & \boldsymbol{t} \end{bmatrix}\begin{bmatrix}\boldsymbol{D}^C_t(\boldsymbol{p}_t)\boldsymbol{K}^{-1}\tilde{\boldsymbol{p}}_t \\ 1\end{bmatrix}}{\boldsymbol{D}^C_s(\boldsymbol{p}_s)} - \tilde{\boldsymbol{p}}_t \\ 
     & = \left( \frac{\boldsymbol{D}^C_t(\boldsymbol{p}_t)}{\boldsymbol{D}^C_s(\boldsymbol{p}_s)} -1 \right) \tilde{\boldsymbol{p}}_t - \frac{\boldsymbol{K} \boldsymbol{t} }{\boldsymbol{D}^C_s(\boldsymbol{p}_s)} 
    \\ 
    & = \frac{t_3}{\boldsymbol{D}^C_s(\boldsymbol{p}_s)} \left( \tilde{\boldsymbol{p}}_t-\tilde{\boldsymbol{p}}_o - {\boldsymbol{K}}\begin{bmatrix}
    t_1/t_3 \\[0.4ex]
    t_2/t_3 \\[0.4ex]
    0
    \end{bmatrix} \right), \\
    \end{aligned}
\end{equation}
where $\boldsymbol{D}^C_s(\boldsymbol{p}_s) = \boldsymbol{D}^C_t(\boldsymbol{p}_t) - t_3$ under the condition of $\boldsymbol{R}=\boldsymbol{I}$, and $\tilde{\boldsymbol{p}}_o$ denotes the homogeneous coordinates of the image principal point. We calculate the divergence of the optical flow at $\boldsymbol{p}_t$ as follows:
\begin{equation}\label{eq.partial_derivatives_of}
    \begin{aligned}
        &\nabla \cdot{}  \boldsymbol{F}^{\rm Tra}_{t\to s}(\boldsymbol{p}_t) 
    = \nabla \cdot{} \left (  \frac{t_3}{\boldsymbol{D}^C_t(\boldsymbol{p}_t)-t_3}\boldsymbol{q}_t \right )
    \\
    =  & \boldsymbol{q}_{t} \cdot{} \nabla \frac{t_3}{\boldsymbol{D}^C_t(\boldsymbol{p}_t)-t_3} 
    +  \frac{t_3}{\boldsymbol{D}^C_t(\boldsymbol{p}_t)-t_3} \nabla \cdot{}
    \boldsymbol{p}_t,
    \end{aligned}
\end{equation}
where $\nabla = [\displaystyle\frac{\partial}{\partial u}, \frac{\partial}{\partial v}]^{\top}$, and $\boldsymbol{q}_t$ is expressed as follows:
\begin{equation}
    \begin{bmatrix}
       \boldsymbol{q}_t \\[0.4ex] 0
    \end{bmatrix}
    =\tilde{\boldsymbol{p}}_t-\tilde{\boldsymbol{p}}_o - \boldsymbol{K}\begin{bmatrix}
    {t_1}/{t_3} \\[0.4ex]
    {t_2}/{t_3} \\[0.4ex]
    0
    \end{bmatrix}.
    \label{eq.qt}
\end{equation}
Rewriting (\ref{eq.partial_derivatives_of}) into the following expression:
\begin{equation}
    \begin{aligned}
        &\underbrace{\frac{\boldsymbol{D}^C_t(\boldsymbol{p}_t)-t_3}{t_3} \nabla \cdot{} \boldsymbol{F}^{\rm Tra}_{t\to s}(\boldsymbol{p}_t) -   \nabla \cdot{} \boldsymbol{p}_{t}}_{\boldsymbol{C}^F(\boldsymbol{p}_t)}  \\
    = &\underbrace{-\frac{\boldsymbol{q}_{t}}{\boldsymbol{D}^C_t(\boldsymbol{p}_t)-t_3 }\cdot{} \nabla \boldsymbol{D}^C_t(\boldsymbol{p}_t)}_{\boldsymbol{C}^D(\boldsymbol{p}_t)}.
    \end{aligned}
\end{equation}
where $\boldsymbol{C}^{F,D} \in \mathbb{R}^{H \times W}$ denotes the differential properties derived from optical flow divergence and depth gradient, respectively. This explicit relationship can be used to provide an additional constraint that helps further improve depth quality from the aspect of local variations via the following DPC loss:
\begin{equation}
    \mathcal{L}_{d} = \frac{1}{HW}  \sum_{ \boldsymbol{p}} \frac{\left |\boldsymbol{C}^D(\boldsymbol{p}) -\boldsymbol{C}^F(\boldsymbol{p}) \right |}{\left |\boldsymbol{C}^D(\boldsymbol{p})\right |}.
    \label{eq.l_d}
\end{equation}
Similar to (\ref{eq.lcgdc}), we consider the relative error in (\ref{eq.l_d}). The effectiveness of our proposed DPC loss is validated through an ablation study detailed in Sect. \ref{section:abl}.

\subsection{Bidirectional Stream Co-Adjustment Strategy}

In the conventional PCG stream, rigid flow is generated using the outputs from DepthNet and PoseNet, which are indirectly supervised by minimizing the photometric loss. DepthNet, when trained in this manner, often struggles in texture-less regions, where a pixel in the target video frame might correspond to multiple pixels with similar intensities in the source frame. Therefore, we are motivated to improve depth estimation in these regions by leveraging the dense correspondences provided by a well-trained FlowNet.

In our proposed CPG stream, despite the effectiveness of infusing dense correspondence priors provided by a pre-trained FlowNet into monocular depth estimation through our developed CGDC and DPC losses, these priors often capture independent apparent motions unrelated to depth when dynamic objects are involved. This can mislead DepthNet, resulting from inaccurately triangulated depths and the misalignment between optical flow divergence and depth gradient on such objects. A straightforward solution to this issue is to exclude dynamic regions using the auto-masking technique developed in \cite{godard2019digging} when computing CGDC and DPC losses. Nonetheless, as illustrated in Fig. \ref{fig.motivation}(c), this technique is not sufficiently robust, as only the dynamic objects that are relatively stationary with respect to ego-motion can be effectively masked \cite{godard2019digging}, and static regions, especially those with low texture, tend to be overly masked \cite{bello2024self}. 

\begin{figure}[t!]
	\centering
	\includegraphics[width=0.49\textwidth]{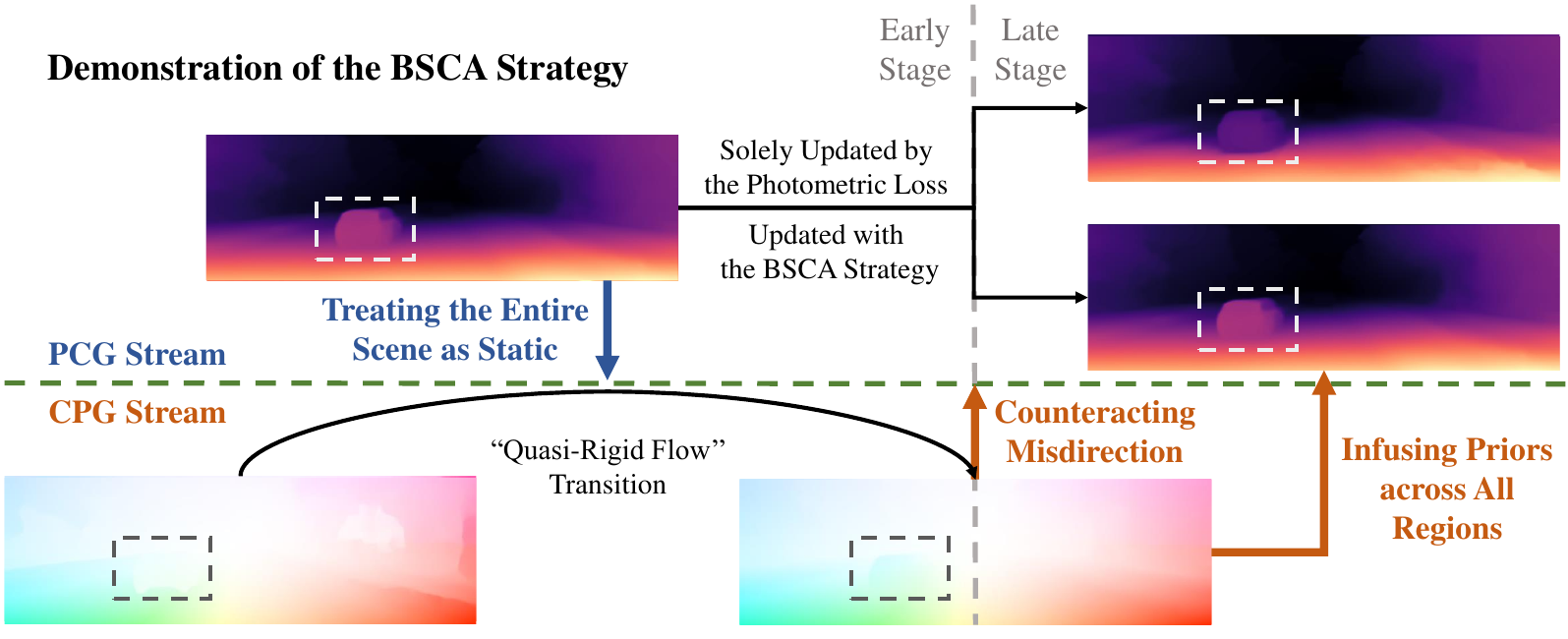}
	\caption{
        An illustration of the interaction between PCG and CPG streams through the proposed BSCA strategy to address the challenges posed by dynamic objects.}
\label{fig:bsca}
\end{figure}

To address the aforementioned two challenges simultaneously, we propose a simple yet effective BSCA strategy, in which the PCG stream and CPG stream complement each other. 
We unfreeze the pre-trained FlowNet during training and jointly optimize the optical flow estimated in the CPG stream and the rigid flow generated in the PCG stream by minimizing the following loss function:
\begin{equation}
\label{eq.co-teaching-loss}
    \mathcal{L}_{b} = \frac{1}{HW}  \sum_{ \boldsymbol{p}} \frac{\left \| \boldsymbol{F}^R_{t \to s}(\boldsymbol{p}) - \boldsymbol{F}_{t \to s}^O(\boldsymbol{p}) \right \| _1}{\left \| \boldsymbol{F}_{t \to s}^O(\boldsymbol{p}) \right \| _1}.
\end{equation} 
In static regions, optical flow and rigid flow should ideally be identical, as demonstrated in prior studies \cite{zou2018df, sun2023dynamo}. 
Similar to \cite{zou2018df}, (\ref{eq.co-teaching-loss}) allows both flows to adjust together without compromising photometric consistency. It particularly encourages rigid flow in the PCG stream towards more accurate correspondences. 
On the other hand, in dynamic regions, optical flow and rigid flow should differ significantly. However, the study \cite{zou2018df} introduces an untrained FlowNet to form a joint learning framework, where both FlowNet and DepthNet are trained by minimizing the photometric loss. Directly minimizing (\ref{eq.co-teaching-loss}) in this training paradigm can cause the DepthNet to be further misled in dynamic regions. 
In contrast, we decouple the FlowNet training from this joint learning framework, generating prior optical flow using a pre-trained FlowNet, which is updated solely by (\ref{eq.co-teaching-loss}). As shown in Fig. \ref{fig:bsca}, by ignoring the photometric consistency constraint on the FlowNet, optical flow in our CPG stream transitions to ``quasi-rigid flow'' under the guidance of static scene hypothesis in the PCG stream at the early stage of training. 
This enables our CGDC and DPC losses in the CPG stream to be effectively applied across the entire image, irrespective of static and dynamic regions.
Moreover, as training progresses, photometric loss tends to mislead DepthNet’s predictions. At this time, (\ref{eq.co-teaching-loss}) counteracts the misdirection caused by the photometric loss, ensuring that the DepthNet performs accurately in dynamic regions.
We conduct an ablation study in Sect. \ref{section:abl} to demonstrate the effectiveness of this strategy.

\begin{table*}[!t]
\begin{center}
\settablefont
\caption{Quantitative comparison with SoTA networks on the KITTI \cite{geiger2012we}, DDAD \cite{guizilini20203d}, nuScenes \cite{caesar2020nuscenes} and Waymo Open \cite{mei2022waymo} benchmarks. The best results are shown in bold symbol. The symbols $\uparrow$ and $\downarrow$ indicate that higher and lower values correspond to better performance, respectively. All models are trained with monocular video sequences. The baseline models used for our experiments on each dataset are underlined, respectively.}
\renewcommand{\arraystretch}{1}
\label{table:compare_with_sota}
{
\begin{tabular}{llcc|cccc|ccc}
\toprule

Dataset & Method & Year & Resolution (pixels) & Abs Rel $\downarrow$ & Sq Rel $\downarrow$ & RMSE $\downarrow$ & RMSE log $\downarrow$ & $\delta <1.25$ $\uparrow$ & $\delta <1.25^2$ $\uparrow$ & $\delta <1.25^3$ $\uparrow$  \\
\hline
\multirow{9}{*}{\makecell[c]{\textbf{KITTI}}} & Monodepth2\cite{godard2019digging} & 2019 & 192 $\times$ 640 & 0.115 & 0.903 & 4.863 & 0.193 & 0.877 & 0.959 & 0.981 \\
 & DIFFNet\cite{zhou2021self} & 2021 & 192 $\times$ 640 & 0.102 & 0.764 & 4.483 & 0.180 & 0.890 & 0.964 & 0.983 \\
 & DaCCN\cite{han2023self} & 2023 & 192 $\times$ 640 & 0.099 & 0.661 & 4.316 & 0.173 & 0.897 & 0.967 & 0.985 \\
 & Dynamo-Depth\cite{sun2023dynamo} & 2023 & 192 $\times$ 640 & 0.112 & 0.758 & 4.505 & 0.183 & 0.873 & 0.959 & 0.984 \\
 & \underline{Lite-Mono-8M}\cite{zhang2023lite} & 2023 & 192 $\times$ 640 & 0.101 & 0.729 & 4.454 & 0.178 & 0.897 & 0.965 & 0.983 \\
 & SENSE\cite{li2023sense} & 2023 & 192 $\times$ 640 & 0.104 & 0.693 & 4.294 & 0.177 & 0.894 & 0.965 & 0.984 \\
 & MonoDiffusion\cite{shao2024monodiffusion} & 2024 & 192 $\times$ 640 & 0.099 & 0.702 & 4.385 & 0.176 & 0.899 & 0.966 & 0.984 \\
 & AQUANet\cite{bello2024self} & 2024 & 192 $\times$ 640 & 0.105 & \textbf{0.621} & \textbf{4.227} & 0.179 & 0.889 & 0.964 & 0.984 \\
\rowcolor{purple!10}  & \textbf{DCPI-Depth (Ours)} & - & 192 $\times$ 640 & \textbf{0.095} &0.662 &4.274 &\textbf{0.170} &\textbf{0.902} &\textbf{0.967} &\textbf{0.985} \\
\hline
\multirow{7}{*}{\makecell[c]{\textbf{DDAD}}} & Monodepth2\cite{godard2019digging} & 2019 & 384 $\times$ 640 & 0.239 & 12.547 & 18.392 & 0.316 & 0.752 & 0.899 & 0.949 \\
& DIFFNet\cite{zhou2021self} & 2021 & 384 $\times$ 640 &   0.205  &  12.126  &  18.461  &   0.289  &   0.795  &   0.916  &   0.957  \\
& SC-Depth\cite{bian2021unsupervised} & 2021 & 384 $\times$ 640 & 0.169 & 3.877 & 16.290 & 0.280 & 0.773 & 0.905 & 0.951 \\
& \underline{Lite-Mono}\cite{zhang2023lite} & 2023 & 384 $\times$ 640 & 0.161  & 4.451  & 16.261  & 0.271  & 0.802  & 0.921  & 0.962 \\
& Lite-Mono-8M\cite{zhang2023lite} & 2023 & 384 $\times$ 640 & 0.175  & 6.425  & 16.687  & 0.272  & 0.799  & 0.920  & 0.961 \\
& SC-DepthV3\cite{sun2023sc} & 2023 & 384 $\times$ 640 & 0.142 & 3.031 & 15.868 & 0.248 & 0.813 & 0.922 & 0.963 \\
\rowcolor{purple!10}  & \textbf{DCPI-Depth (Ours)} & - & 384 $\times$ 640 & \textbf{0.140} & \textbf{2.866} & \textbf{15.786} & \textbf{0.238} & \textbf{0.815} & \textbf{0.929} & \textbf{0.970} \\

\hline
\multirow{7}{*}{\makecell[c]{\textbf{nuScenes}}} & Monodepth2\cite{godard2019digging} & 2019 & 288 $\times$ 512 & 0.425 & 16.592 & 10.040 & 0.402 & 0.723 & 0.837 & 0.887 \\
 & DIFFNet\cite{zhou2021self} & 2021 & 288 $\times$ 512 & 0.228 & 5.925 & 8.897 & 0.290 & 0.772 & 0.905 & 0.950 \\ 
 & MonoViT-tiny\cite{zhao2022monovit} & 2022 & 288 $\times$ 512 & 0.412 & 16.061 & 10.504 & 0.385 & 0.717 & 0.842 & 0.898 \\
 & \underline{Lite-Mono}\cite{zhang2023lite} & 2023 & 288 $\times$ 512 & 0.419 & 15.578 & 9.807 & 0.449 & 0.720 & 0.831 & 0.879 \\
 & Lite-Mono-8M\cite{zhang2023lite} & 2023 & 288 $\times$ 512 & 0.429 & 17.058 & 10.559 & 0.400 & 0.709 & 0.830 & 0.883 \\
 & Dynamo-Depth\cite{sun2023dynamo} & 2023 & 288 $\times$ 512 & 0.179 & 2.118 & \textbf{7.050} & 0.271 & 0.787 & 0.896 & 0.940 \\
\rowcolor{purple!10}  & \textbf{DCPI-Depth (Ours)} & - & 288 $\times$ 512 & \textbf{0.160} & \textbf{1.736} & 7.194 & \textbf{0.248} & \textbf{0.793} & \textbf{0.921} & \textbf{0.966} \\

\hline
\multirow{7}{*}{\makecell[c]{\textbf{Waymo}}} & Monodepth2\cite{godard2019digging} & 2019 & 320 $\times$ 480 & 0.173 & 2.731 & 7.708 & 0.227 & 0.797 & 0.930 & 0.968 \\
 & DIFFNet\cite{zhou2021self} & 2021 & 320 $\times$ 480 & 0.149 & 2.082 & 7.474 & 0.200 & 0.838 & 0.956 & 0.981 \\
 & Li \textit{et al.}\cite{li2021unsupervised} & 2021 & 320 $\times$ 480 & 0.157 & 1.531 & 7.090 & 0.205 & -- & -- & -- \\
 & \underline{Lite-Mono}\cite{zhang2023lite} & 2023 & 320 $\times$ 480 & 0.158 & 2.305 & 7.394 & 0.210 & 0.816 & 0.944 & 0.976 \\
 & Lite-Mono-8M\cite{zhang2023lite} & 2023 & 320 $\times$ 480 & 0.154 & 2.297 & 7.495 & 0.209 & 0.825 & 0.947 & 0.975 \\
 & Dynamo-Depth\cite{sun2023dynamo} & 2023 & 320 $\times$ 480 & \textbf{0.116} & 1.156 & 6.000 & 0.166 & \textbf{0.878} & 0.969 & 0.989 \\
\rowcolor{purple!10}  & \textbf{DCPI-Depth (Ours)} & - & 320 $\times$ 480 & \textbf{0.116} & \textbf{0.963} & \textbf{5.642} & \textbf{0.162} & 0.872 & \textbf{0.972} & \textbf{0.991} \\

\bottomrule
\end{tabular}
}
\end{center}
\end{table*}

\section{Experiments}
\label{Sect.experiments}

The performance of our proposed DCPI-Depth is evaluated both qualitatively and quantitatively with extensive experiments in this section. The following subsections provide details on the utilized datasets, practical implementation, evaluation metrics, ablation studies, and comprehensive comparisons with other SoTA methods.

\subsection{Datasets}
\label{sec:dataset}

We conduct our experiments on six public datasets: \textbf{KITTI}\cite{geiger2012we}, \textbf{DDAD}\cite{guizilini20203d}, \textbf{nuScenes} \cite{caesar2020nuscenes}, \textbf{Waymo} Open \cite{mei2022waymo}, \textbf{Make3D}\cite{saxena2008make3d}, and \textbf{DIML} \cite{cho2021diml}. 

For the \textbf{KITTI}\cite{geiger2012we} dataset, we adopt the Eigen split \cite{eigen2014depth}, which comprises 39,180 monocular triplets for training, 4,424 images for validation, and 697 images for testing. For the \textbf{DDAD}\cite{guizilini20203d} dataset, we follow the prior work \cite{sun2023sc} to split this dataset into a training set of 12,650 images and a test set of 3,950 images in our experiments. For the \textbf{nuScenes} \cite{caesar2020nuscenes} dataset, following \cite{sun2023dynamo}, we use 79,760 image triplets collected by the front camera for model training, and evaluate the model's performance on 6,019 front camera images. For the \textbf{Waymo} Open \cite{mei2022waymo} dataset, as in \cite{sun2023dynamo}, we utilize 76,852 front camera image triplets for training, and 2,216 front camera images for evaluation. For the \textbf{Make3D}\cite{saxena2008make3d} and \textbf{DIML} \cite{cho2021diml} datasets, since neither stereo image pairs nor monocular sequences are provided for unsupervised training, we only use this dataset to quantify the generalizability of models per-trained on the KITTI dataset.

\subsection{Experimental Setup}
\label{section:Experimental Setup}
Our experiments are conducted on an NVIDIA RTX 4090
GPU with a batch size of 12. 
Following \cite{godard2019digging}, we adopt a training approach wherein a snippet of three consecutive video frames is utilized as a training sample. To augment the dataset, random color jitter and horizontal flips are applied to the images during model training.
To minimize the loss functions, we employ the AdamW optimizer with an initial learning rate of $1 \times 10^{-4}$ and a weight decay of $1 \times 10^{-2}$. The learning rate is adjusted using a cosine annealing scheduler with periodic restarts, decaying from $1 \times 10^{-4}$ to $5 \times 10^{-6}$ over 31 epochs. A decay factor of $\gamma = 0.9$ is applied after each cycle to ensure gradual reduction in learning rates.
Following \cite{godard2019digging, zhang2023lite}, the network's encoder is initialized using pre-trained weights from the ImageNet database \cite{deng2009imagenet}.
RAFT \cite{teed2020raft} is employed as the FlowNet in our framework to provide dense correspondence prior. It is pre-trained on the KITTI Flow 2015 \cite{menze2015object} dataset, which contains only 200 sets of two consecutive frames and has a small overlap with the KITTI Eigen split \cite{eigen2014depth}.

\subsection{Evaluation Metrics}
We employ seven metrics to quantify the model’s performance: mean absolute relative error (Abs Rel), mean squared relative error (Sq Rel), root mean squared error (RMSE), root mean squared log error (RMSE log), and the accuracy under specific thresholds ($\delta_i < 1.25^i$, where $i = 1, 2, 3$). Detailed expressions for these metrics can be found in \cite{eigen2014depth}.

\begin{figure*}[t!]
	\centering
	\includegraphics[width=0.98 \textwidth]{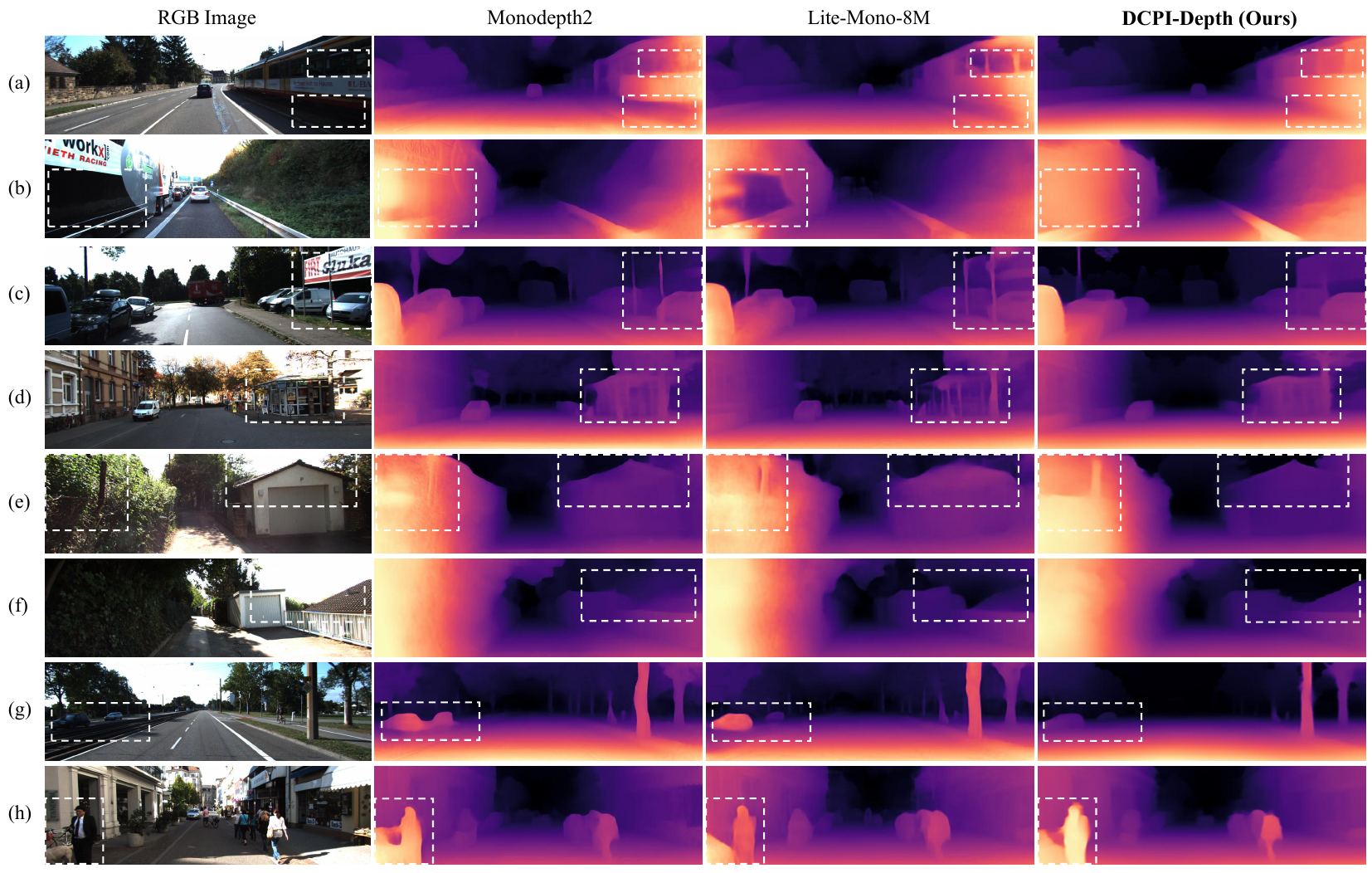}
	\caption{
        Qualitative comparisons among Monodepth2, Lite-Mono-8M, and our proposed DCPI-Depth on the KITTI \cite{geiger2012we} dataset. (a)-(b), (c)-(d), (e)-(f), and (g)-(h) demonstrate the robustness of DCPI-Depth in texture-less regions, in texture-rich regions, at static object boundaries, and on dynamic objects, respectively. 
        }
\label{fig.depth_compare_sota}
\end{figure*}

\begin{figure*}[t!]
	\centering
	\includegraphics[width=0.99 \textwidth]{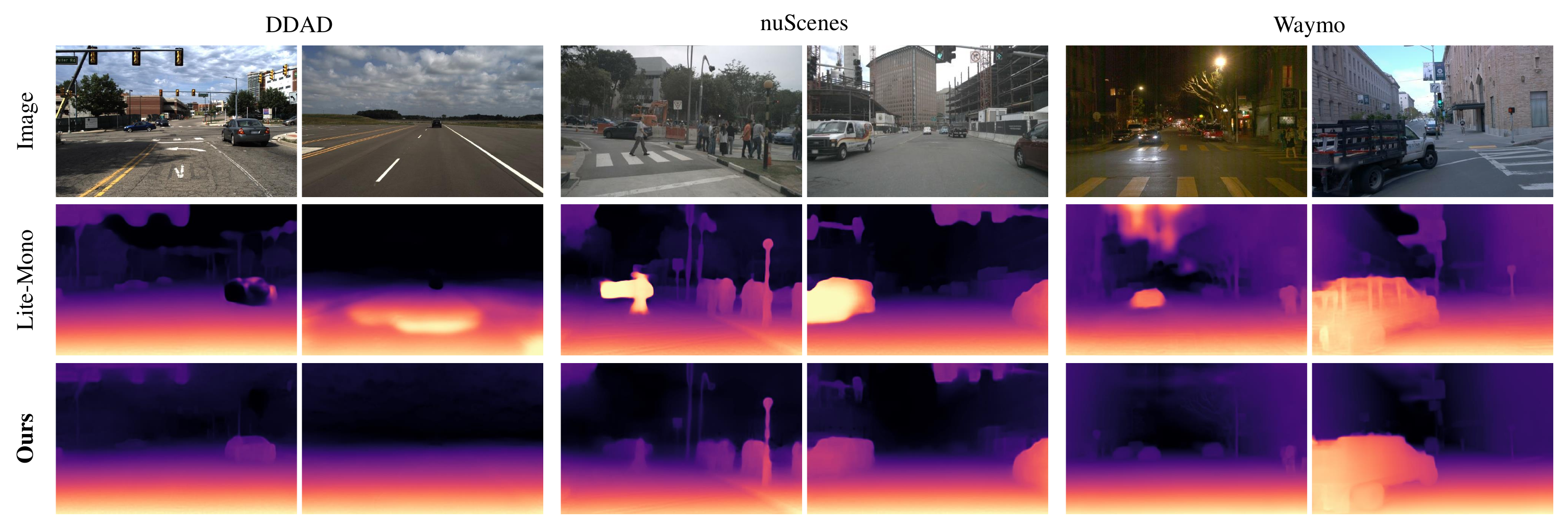}
	\caption{Qualitative comparisons between Lite-Mono and our proposed DCPI-Depth on the DDAD                       \cite{guizilini20203d}, nuScenes \cite{caesar2020nuscenes} and Waymo Open \cite{mei2022waymo} datasets. }
\label{fig.depth_compare_dnw}
\end{figure*}

\begin{table*}[!t]
\begin{center}
			\settablefont
			\caption{Quantitative comparison with SoTA networks on the KITTI \cite{geiger2012we} dataset using the improved KITTI ground truth provided in \cite{uhrig2017sparsity} for testing or higher-resolution images for training. The best results are shown in bold symbol. The symbols $\uparrow$ and $\downarrow$ indicate that higher and lower values correspond to better performance, respectively. ``M'' denotes training with monocular video sequences, and ``MS'' denotes training with both monocular video sequences and stereo image pairs. $^\dagger$ indicates the results achieved using the same weights in Table \ref{table:compare_with_sota} for consistency.
            }
			\label{table:kitti_imp}
			\begin{tabular}{lccc|cccc|ccc}
				\toprule
				Method & Year & Resolution (pixels) & Data & Abs Rel $\downarrow$ & Sq Rel $\downarrow$ & RMSE $\downarrow$ & RMSE log $\downarrow$ & $\delta <1.25$ $\uparrow$&  $\delta <1.25^2$ $\uparrow$ & $\delta <1.25^3$ $\uparrow$ \\
                
				\hline
				Monodepth2 \cite{godard2019digging}		& 2019	& 192 $\times$ 640 & MS & 0.080 & 0.466 & 3.681 & 0.127 & 0.926 & 0.985 & 0.995 \\
                DIFFNet$^\dagger$ \cite{zhou2021self} & 2021 & 192 $\times$ 640 &M &   0.076  &   0.414  &   3.492  &   0.119  &   0.936  &   0.988  &   0.996  \\
                Lite-Mono$^\dagger$ \cite{zhang2023lite}             & 2023  & 192 $\times$ 640 & M 	    &   0.077  &   0.413  &   3.482  &   0.119  &   0.933  &   0.988  &   0.997  \\
                Lite-Mono-8M$^\dagger$ \cite{zhang2023lite}             & 2023  & 192 $\times$ 640 & M 	&   0.077  &   0.423  &   3.527  &   0.119  &   0.934  &   0.988  &   0.997  \\
                SENSE\cite{li2023sense} & 2023 & 192 $\times$ 640 &M & 0.071 & 0.339 & 3.175 & 0.109 & 0.945 & 0.990 & \textbf{0.998} \\
                MonoDiffusion\cite{shao2024monodiffusion} & 2024 & 192 $\times$ 640 &M & 0.073 & 0.377 & 3.451 & 0.115 & 0.935 & 0.988 & 0.997 \\
                AQUANet\cite{bello2024self} & 2024 & 192 $\times$ 640 &M & 0.070 & \textbf{0.285} & \textbf{2.988} & \textbf{0.107} & 0.948 & \textbf{0.992} & \textbf{0.998} \\
				\rowcolor{purple!10}\textbf{DCPI-Depth$^\dagger$ (Ours)}  & - 		& 192 $\times$ 640 & M  &   \textbf{0.066}  &   0.326  &   3.257  &   \textbf{0.107}  &   \textbf{0.949}  &   0.990  &   0.997  \\

                \hline
                Monodepth2\cite{godard2019digging} & 2019 & 320 $\times$ 1024 &M & 0.115 & 0.882 & 4.701 & 0.190 & 0.879 & 0.961 & 0.982 \\
                DIFFNet\cite{zhou2021self} & 2021 & 320 $\times$ 1024 &M & 0.097 & 0.722 & 4.435 & 0.174 & 0.907 & 0.967 & 0.984 \\
                MonoViT-tiny\cite{zhao2022monovit} & 2022 & 320 $\times$ 1024 &M & 0.096 & 0.714 & 4.292 & 0.172 & 0.908 & 0.968 & 0.984 \\
                DaCCN\cite{han2023self} & 2023 & 320 $\times$ 1024 &M & 0.094 & 0.624 & 4.145 & 0.169 & 0.909 & \textbf{0.970} & \textbf{0.985} \\
                SENSE\cite{li2023sense} & 2023 & 320 $\times$ 1024 &M & 0.099 & \textbf{0.617} & \textbf{4.079} & 0.172 & 0.902 & 0.968 & \textbf{0.985} \\
                MonoDiffusion\cite{shao2024monodiffusion} & 2024 & 320 $\times$ 1024 &M & 0.095 & 0.670 & 4.219 & 0.171 & 0.909 & 0.968 & 0.984 \\
                \rowcolor{purple!10}\textbf{DCPI-Depth (Ours)} & - & 320 $\times$ 1024 &M & \textbf{0.090} &0.655 &4.113 &\textbf{0.167} &\textbf{0.914} &0.969 & \textbf{0.985} \\
                
				\bottomrule
			\end{tabular}
\end{center}
\end{table*}

\subsection{Comparison with SoTA Approaches}

The quantitative experimental results presented in Table \ref{table:compare_with_sota} demonstrate that DCPI-Depth achieves SoTA performance across the KITTI \cite{geiger2012we}, DDAD \cite{guizilini20203d}, nuScenes \cite{caesar2020nuscenes}, and Waymo Open \cite{mei2022waymo} datasets.
Notably, the employed baseline model, Lite-mono \cite{zhang2023lite}, that performs unsatisfactorily on these datasets, achieves significant performance improvements, with error reductions ranging from 13.04\% to 61.81\% on Abs Rel. This substantial enhancement enables DCPI-Depth to surpass previously leading methods \cite{sun2023sc, sun2023dynamo} on each of the respective datasets, suggesting the effectiveness of our proposed framework.

The qualitative experimental results on the KITTI dataset are shown in Fig. \ref{fig.depth_compare_sota}. Our DCPI-Depth exhibits superior performance compared to previous SoTA methods. This is particularly evident in texture-less regions, such as (a) and (b). Additionally, our approach ensures smoother and more continuous depth changes and generates clearer boundaries, as exemplified in (c) to (f). These improvements are attributed to the dense correspondence priors infused via our proposed CGDC and DPC losses within the CPG stream, which provide refined and direct geometric cues for depth estimation. Furthermore, DCPI-Depth maintains accurate depth estimation on dynamic objects, such as (g) and (h), unaffected by independent motion. This advantage, observed on dynamic objects, stems from our BSCA strategy, which not only prevents the misleading effects of independent motion captured by the optical flow within the CPG stream but also reinforces accurate estimations early in the training phase, preserving them through to the final results. 
As shown in Fig. \ref{fig.depth_compare_dnw}, the baseline model performs poorly on the DDAD, nuScenes, and Waymo datasets, but demonstrates dramatic performance improvements after trained under our proposed framework, further demonstrating the robustness of DCPI-Depth across diverse scenarios. 


To further investigate the impact of image resolutions and ground truth quality, We evaluate our network's performance using (1) images at a resolution of 320$\times$1024 pixels and (2) the improved ground truth \cite{uhrig2017sparsity} from the KITTI dataset.
It can be observed in Table \ref{table:kitti_imp} that DCPI-Depth consistently achieves SoTA performances compared with existing methods. 
Although certain metrics exhibit suboptimal results, potentially attributable to the inherent limitation of the network backbone, the proposed training framework, infused with dense correspondence priors, demonstrates a marked improvement in baseline performance and robustness.

\begin{table*}[!t]
\begin{center}
\settablefont
\caption{Quantitative results on the KITTI \cite{geiger2012we} dataset for the full image, static regions, and dynamic regions. The best results are shown in bold symbol. The symbols $\uparrow$ and $\downarrow$ indicate that higher and lower values correspond to better performance, respectively. ``M'' denotes training with monocular video sequences.}
\renewcommand{\arraystretch}{1}
\label{table:compare_with_sc}
{
\begin{tabular}{c|cccc|cccc|ccc}
\toprule

{Region} & Method & Year & Resolution (pixels) & Data & Abs Rel $\downarrow$ & Sq Rel $\downarrow$ & RMSE $\downarrow$ & RMSE log $\downarrow$ & \textbf{$\delta <1.25$} $\uparrow$ & \textbf{$\delta <1.25^2$} $\uparrow$ & \textbf{$\delta <1.25^3$} $\uparrow$  \\
\hline
\multicolumn{1}{c|}{\multirow{4}{*}{{Full Image}}} & SC-Depth\cite{bian2021unsupervised} & 2021 & 256 $\times$ 832 & M & 0.118 & 0.870 & 4.997 & 0.197 & 0.860 & 0.956 & 0.981 \\
\multicolumn{1}{c|}{} & SC-DepthV2\cite{bian2021auto} & 2022 & 256 $\times$ 832 & M & 0.118 & 0.861 & 4.803 & 0.193 & 0.866 & 0.958 & 0.981 \\
\multicolumn{1}{c|}{} & SC-DepthV3\cite{sun2023sc} & 2023 & 256 $\times$ 832 & M & 0.118 & 0.756 & 4.709 & 0.188 & 0.864 & 0.960 & 0.984 \\
\cline{2-12}
\multicolumn{1}{c|}{} &  \cellcolor{purple!10}\textbf{ DCPI-Depth (Ours)} & \cellcolor{purple!10}- & \cellcolor{purple!10}256 $\times$ 832 & \cellcolor{purple!10}M & \cellcolor{purple!10}\textbf{0.109} & \cellcolor{purple!10}\textbf{0.679} & \cellcolor{purple!10}\textbf{4.496} & \cellcolor{purple!10}\textbf{0.180} & \cellcolor{purple!10}\textbf{0.878} & \cellcolor{purple!10}\textbf{0.965} & \cellcolor{purple!10}\textbf{0.985} \\

\hline

\multicolumn{1}{c|}{\multirow{3}{*}{\makecell[c]{Static\\ Regions}}} & SC-Depth\cite{bian2021unsupervised} & 2021 & 256 $\times$ 832 & M & 0.106 & 0.704 & 4.702 & 0.170 & 0.874 & 0.966 & 0.989 \\
\multicolumn{1}{c|}{} & SC-DepthV3\cite{sun2023sc} & 2023 & 256 $\times$ 832 & M & 0.108 & 0.636 & 4.438 & 0.163 & 0.881 & 0.971 & 0.991 \\
\cline{2-12}
\multicolumn{1}{c|}{} & \cellcolor{purple!10}\textbf{DCPI-Depth (Ours)} & \cellcolor{purple!10}- & \cellcolor{purple!10}256 $\times$ 832 & \cellcolor{purple!10}M & \cellcolor{purple!10}\textbf{0.101} & \cellcolor{purple!10}\textbf{0.584} & \cellcolor{purple!10}\textbf{4.235} & \cellcolor{purple!10}\textbf{0.156} & \cellcolor{purple!10}\textbf{0.892} & \cellcolor{purple!10}\textbf{0.974} & \cellcolor{purple!10}\textbf{0.991} \\

\hline
\multicolumn{1}{c|}{\multirow{3}{*}{\makecell[c]{Dynamic\\ Regions}}} & SC-Depth\cite{bian2021unsupervised} & 2021 & 256 $\times$ 832 & M & 0.243 & 3.890 & 8.533 & 0.321 & 0.689 & 0.849 & 0.921 \\
\multicolumn{1}{c|}{} & SC-DepthV3\cite{sun2023sc} & 2023 & 256 $\times$ 832 & M & 0.205 & 2.283 & 7.356 & 0.290 & 0.703 & 0.884 & 0.945 \\
\cline{2-12}
\multicolumn{1}{c|}{} & \cellcolor{purple!10}\textbf{DCPI-Depth (Ours)} & \cellcolor{purple!10}- & \cellcolor{purple!10}256 $\times$ 832 & \cellcolor{purple!10}M & \cellcolor{purple!10}\textbf{0.186} & \cellcolor{purple!10}\textbf{1.948} & \cellcolor{purple!10}\textbf{7.028} & \cellcolor{purple!10}\textbf{0.281} & \cellcolor{purple!10}\textbf{0.732} & \cellcolor{purple!10}\textbf{0.895} & \cellcolor{purple!10}\textbf{0.950} \\

\bottomrule
\end{tabular}
}
\end{center}
\end{table*}

\begin{figure*}[t!]
	\centering
	\includegraphics[width=0.90\textwidth]{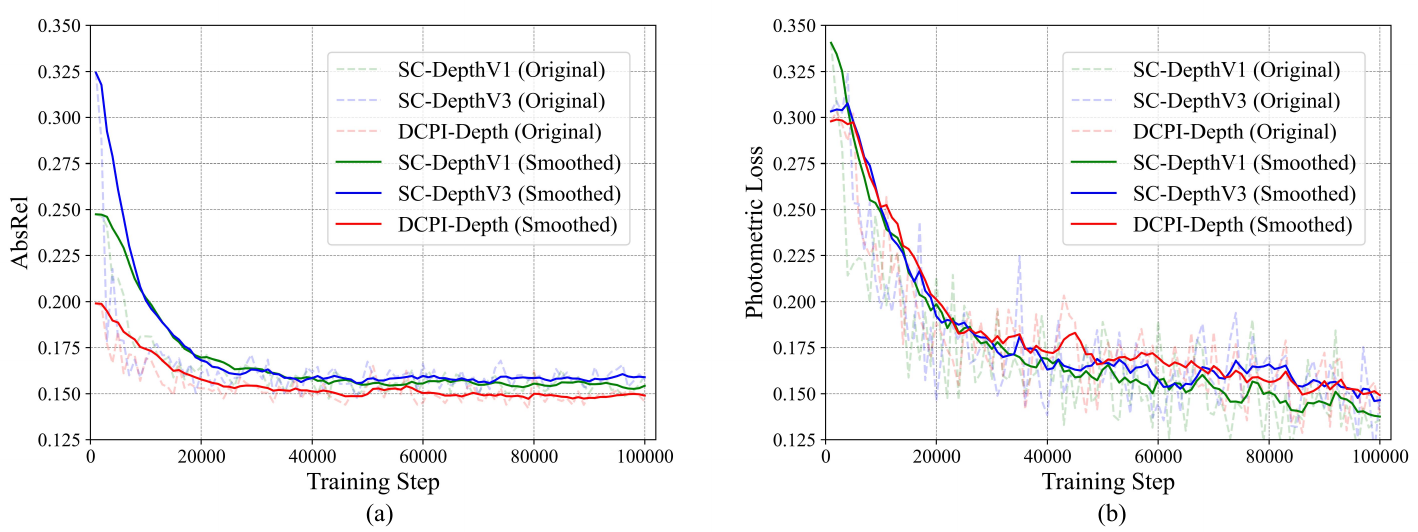}
	\caption{
        Learning curve comparisons among SC-DepthV1, SC-DepthV3, and our proposed DCPI-Depth on the KITTI \cite{geiger2012we} dataset: (a) demonstrates that our DCPI-Depth consistently achieves a lower Abs Rel throughout training compared to other models; (b) illustrates that the convergence of photometric loss among the three models is comparable.}
\label{fig:convergence}
\end{figure*}

\begin{table*}[!t]
\begin{center}
\settablefont
\caption{Ablation studies on the KITTI \cite{geiger2012we} and the DDAD \cite{guizilini20203d} datasets. The best results are shown in bold symbol. The symbols $\uparrow$ and $\downarrow$ indicate that higher and lower values correspond to better performance, respectively.}
\renewcommand{\arraystretch}{1.1}
\label{table:abl}
{
\begin{tabular}{l|ccc|cccc|ccc|cccc}
\toprule
 \multirow{2}{*}{\makecell[c]{DepthNet\\Backbone}} & \multicolumn{2}{c|}{CPG Stream} & \multirow{2}{*}{BSCA Strategy} & \multicolumn{4}{c|}{KITTI (Full Image)} & \multicolumn{3}{c|}{KITTI (Dynamic Regions)} & \multicolumn{4}{c}{DDAD (Full Image)}\\
 \cline{2-3} \cline{5-15}
  & CGDC loss & \multicolumn{1}{c|}{DPC loss} &  & Abs Rel $\downarrow$ & Sq Rel $\downarrow$ & $\delta_1$ $\uparrow$ & $\delta_2$ $\uparrow$ & Abs Rel $\downarrow$ & Sq Rel $\downarrow$ & $\delta_1$ $\uparrow$ & Abs Rel $\downarrow$ & Sq Rel $\downarrow$ & $\delta_1$ $\uparrow$ & $\delta_2$ $\uparrow$ \\

\hline
\multirow{5}{*}{ResNet18 \cite{he2016deep}} & \multicolumn{3}{c|}{Baseline} & 0.118 & 0.756 & 0.864 & 0.960 & 0.205 & 2.283 & 0.703 & 0.149 & 3.062 & 0.798 & 0.920 \\
\cline{2-4}
 & \checkmark & & \checkmark & 0.111 & 0.698 & 0.874 & 0.963 & 0.193 & 2.125 & 0.718 & 0.143 & 3.156 & 0.801 & 0.915 \\
 & & \checkmark & \checkmark & 0.115 & 0.716 & 0.869 & 0.963 & 0.201 & 2.234 & 0.710 & 0.148 & 3.092 & 0.799 & 0.921 \\
 & \checkmark & \checkmark & & 0.113 & 0.707 & 0.873 & 0.964 & 0.211 & 2.510 & 0.700 & 0.145 & 3.066 & 0.805 & 0.919 \\
\cline{2-4}
 & \checkmark & \checkmark & \checkmark & \textbf{0.109} & \textbf{0.679} & \textbf{0.878} & \textbf{0.965} & \textbf{0.186} & \textbf{1.948} & \textbf{0.732} & \textbf{0.143} & \textbf{2.963} & \textbf{0.812} & \textbf{0.925} \\
 \hline
\multirow{2}{*}{DIFFNet \cite{zhou2021self}} & \multicolumn{3}{c|}{Baseline} & 0.108 & 0.696 & 0.878 & 0.964 & 0.197 & 2.188 & 0.717 & 0.145 & 3.104 & 0.804 & 0.922 \\
\cline{2-4}
 & \checkmark & \checkmark & \checkmark & \textbf{0.103} & \textbf{0.653} & \textbf{0.886} & \textbf{0.966} & \textbf{0.191} & \textbf{2.098} & \textbf{0.719} & \textbf{0.137} & \textbf{2.812} & \textbf{0.816} & \textbf{0.926} \\

\bottomrule
\end{tabular}
}
\end{center}
\end{table*}

\begin{figure*}[!t]
	\centering
	\includegraphics[width=0.99\linewidth]{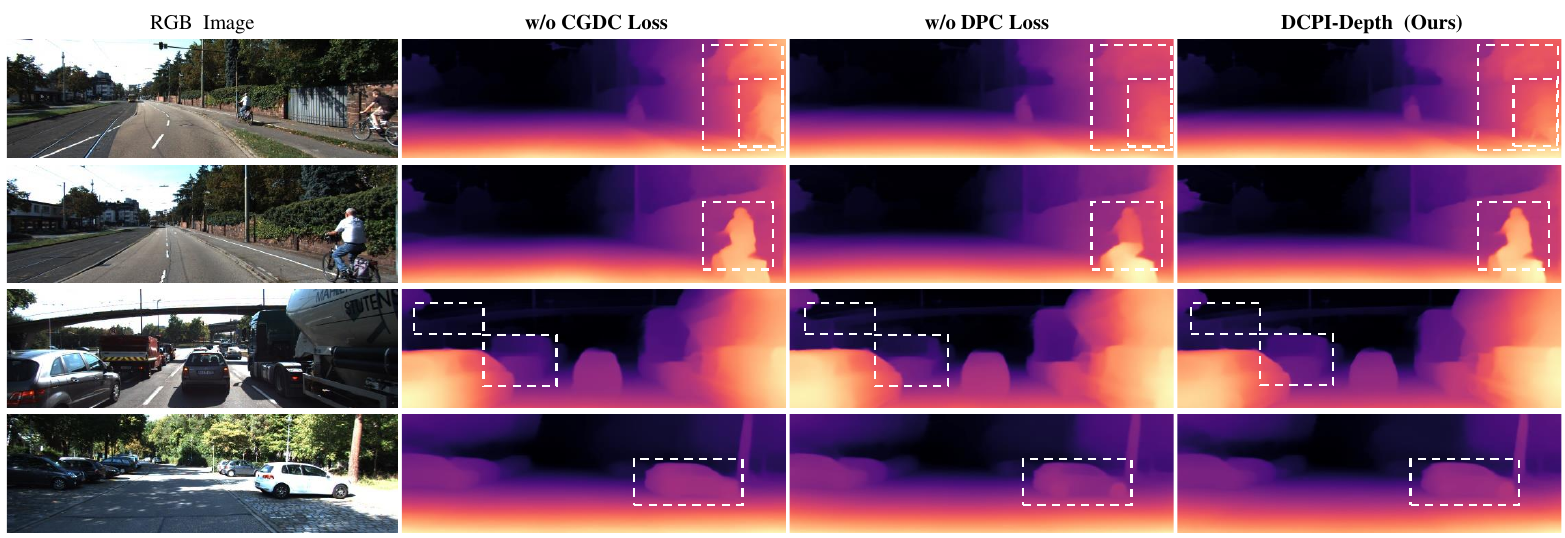}
	\caption{Qualitative results on the KITTI \cite{geiger2012we} dataset to demonstrate the effectiveness of the CGDC loss and DPC loss.}
\label{fig.abl}
\end{figure*}

\begin{figure*}[!t]
	\centering
	\includegraphics[width=0.99\linewidth]{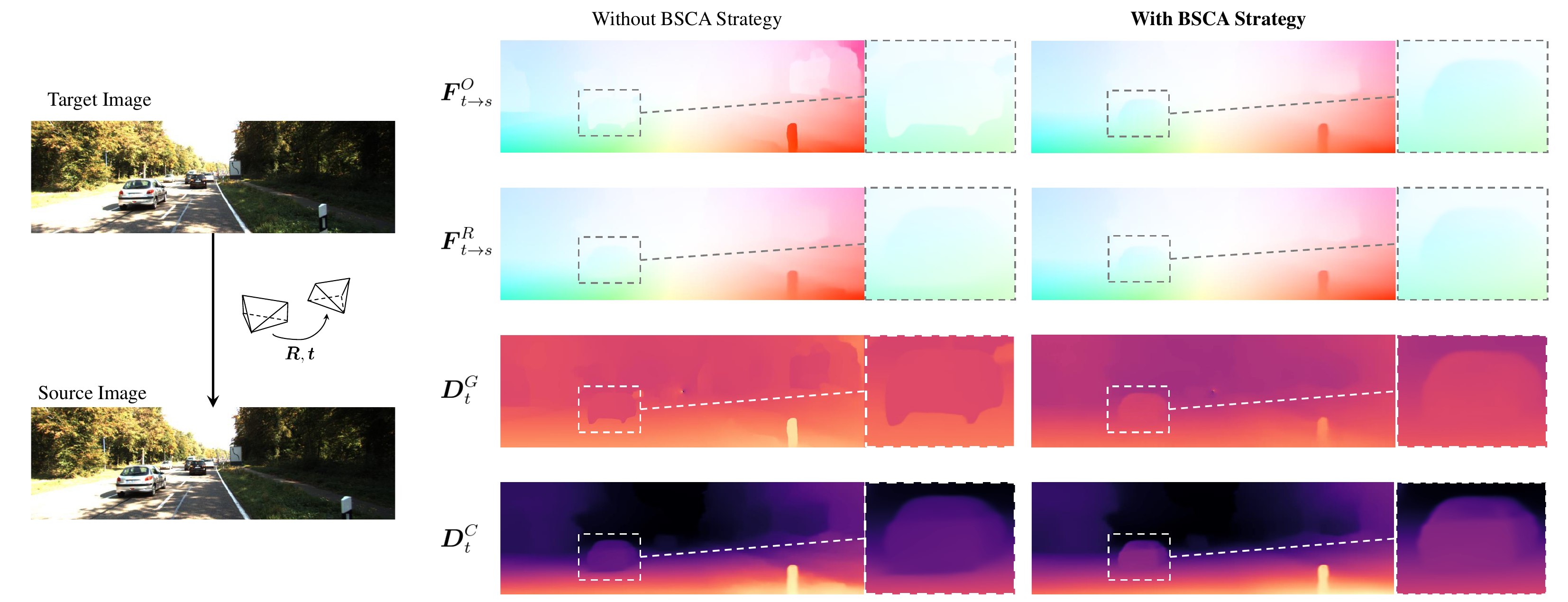}
	\caption{Qualitative results on the KITTI \cite{geiger2012we} dataset to demonstrate the effectiveness of the BSCA strategy.}
\label{fig.flowabl}
\end{figure*}

To further validate the effectiveness of DCPI-Depth in leveraging dense correspondence priors, we deploy our framework to SC-DepthV3 \cite{sun2023sc}, which leverages pseudo depth to achieve robust and highly reliable depth estimation. It also provides a comprehensive platform for evaluating depth performance across dynamic objects, static areas, and full images. As shown in Table \ref{table:compare_with_sc}, DCPI-Depth achieves significant improvements, outperforming the baseline by a considerable margin across all metrics for full image, static regions, and dynamic regions. Furthermore, we provide comparisons of learning curves among SC-DepthV1, SC-DepthV3, and DCPI-Depth. 
As illustrated in Fig. \ref{fig:convergence}(a), DCPI-Depth significantly improves depth estimation performance compared to SC-DepthV3. 
These results demonstrate that the dense correspondence prior we incorporate additionally provides meaningful geometric cues on top of pseudo depth. Moreover, it can be observed in Fig. \ref{fig:convergence}(b) that the convergence of the photometric loss is comparable among the three models. This observation suggests that merely minimizing photometric loss, which provides an indirect supervisory signal, presents challenges in further improving depth estimation performance. In contrast, our approach provides a more direct and effective constraint for depth estimation. 

\subsection{Ablation Studies}
\label{section:abl}

Table \ref{table:abl} presents comprehensive ablation studies conducted with SC-DepthV3 to validate the effectiveness of our contributed components. The first ablation study validates the internal design of the CPG stream by comparing the overall performance with and without the incorporation of the CGDC loss and DPC loss, respectively. Key findings from this study include:
(1) Incorporating the CGDC loss results in the most significant performance improvements. 
(2) The DPC loss also plays an important role in improving the performance of our framework. Without the DPC loss, the Abs Rel metric, which measures the relative depth error similar to the CGDC loss, remains steady. However, other metrics decrease significantly, especially on the DDAD dataset.
Fig. \ref{fig.abl} provides quantitative experimental results to demonstrate the effectiveness of the CGDC and DPC losses. It is obvious that removing the CGDC loss can lead to significant depth distribution shifts (see the first and third lines), which aligns with our expectations. Furthermore, despite achieving satisfactory quantitative results, removing the DPC loss can result in erroneous depth estimations in continuous regions such as vehicles and cyclists, as illustrated in the second to fourth lines.


Additionally, we conduct another ablation study where we omit the BSCA strategy while retaining the full configuration of the CPG stream to validate its efficacy. The quantitative results reveal a significant decline in performance in dynamic regions, even falling below that of the baseline network, while the depth estimation accuracy in static regions remains unaffected. The qualitative results are provided in Fig. \ref{fig.flowabl}. It can be observed that the moving car exhibits independent apparent motions in the optical flow provided by the frozen FlowNet, leading to erroneous depth guidance via triangulation. As a result, the rigid flow eventually tends to contain the independent apparent motions, and the estimated depth is farther than the actual. In contrast, the model trained with the BSCA strategy effectively eliminates independent apparent motions in both flows for the moving car. This results in more accurate geometric and contextual-based depth estimations for the moving car. The above observations are consistent with our initial motivation for introducing the BSCA strategy, emphasizing its critical role in improving depth estimation performance when dynamic objects are involved. 

Thirdly, we conduct an ablation study using a better-performing DepthNet backbone within our full configuration. The results show that our contributed techniques are compatible with this backbone and consistently achieve significant improvements. These findings indicate that our contributions provide distinct advantages that differentiate them from those provided by more advanced networks and demonstrate the potential to deliver improvements across a wider range of models.


\begin{table}[!t]
\begin{center}
\settablefont
\caption{Quantitative experimental results for additional monocular depth estimation models, with and without our proposed DCPI-Depth framework employed, using the KITTI Eigen split \cite{eigen2014depth}. All models are trained on 192$\times$640 images. }
\label{table:Comparisons_of_existing_models}
\renewcommand{\arraystretch}{1.1}
{
\begin{tabular*}{\linewidth}{@{\hspace{7pt}}c@{\hspace{8pt}}|c@{\hspace{9pt}}c@{\hspace{9pt}}c@{\hspace{9pt}}c|c@{\hspace{10pt}}c@{\hspace{10pt}}c}
\toprule
Method & Abs Rel & Sq Rel & RMSE & RMSE log & $\delta_1$ & $\delta_2$ & $\delta_3$  \\
\hline
Monodepth2\cite{godard2019digging} &  0.115 & 0.903 & 4.863 & 0.193 & 0.877 & 0.959 & 0.981 \\
\textbf{ +Ours} & \textbf{0.110} &\textbf{0.749} &\textbf{4.559} &\textbf{0.183}  &\textbf{0.879} &\textbf{0.962} &\textbf{0.984} \\
\hline
Swin-Depth\cite{shim2023swindepth} &  0.106 & 0.739 & 4.510 & 0.182 & 0.890 & 0.964 & 0.984 \\
\textbf{ +Ours} & \textbf{0.104} &\textbf{0.704} &\textbf{4.443} &\textbf{0.180} &\textbf{0.893} &\textbf{0.965} &0.984 \\
\hline
DIFFNet\cite{zhou2021self} &  0.102 & 0.764 & 4.483 & 0.180 & 0.896 & 0.965 & 0.983 \\
\textbf{ +Ours} & \textbf{0.096} &\textbf{0.670} &\textbf{4.310} &\textbf{0.172} &\textbf{0.901} &\textbf{0.966} &\textbf{0.984} \\
\hline
Lite-Mono-3M\cite{zhang2023lite} &  0.107 & 0.765 & 4.561 & 0.183 & 0.886 & 0.963 & 0.983 \\
\textbf{ +Ours} & \textbf{0.101} &\textbf{0.707} &\textbf{4.447} &\textbf{0.176} &\textbf{0.893} &\textbf{0.964} &\textbf{0.984} \\

\bottomrule
\end{tabular*}
}
\end{center}
\end{table}

	\begin{table}[!t]
		\begin{center}
			\settablefont
			\caption{Quantitative results on the Make3D \cite{saxena2008make3d} and DIML \cite{cho2021diml} datasets. All models are trained on the KITTI \cite{geiger2012we} dataset images (resolution: 640$\times$192 pixels).}
			\label{table:make3d}
			\begin{tabular}{ll|cccc}
				\toprule
				Dataset & Method     & Abs Rel & Sq Rel & RMSE  & RMSE log \\ \hline
				\multirow{6}{*}{\makecell[c]{\textbf{Make3D}}} &Monodepth2 \cite{godard2019digging} 	& 0.321   & 3.378  & 7.252 & 0.163    \\
				& HR-Depth \cite{lyu2021hr}  		& 0.305   & 2.944  & 6.857 & 0.157    \\
                & R-MSFM6 \cite{yan2021channel}   		& 0.334   & 3.285  & 7.212 & 0.169    \\

				& Lite-Mono \cite{zhang2023lite}   		& 0.305   & 3.060  & 6.981 & 0.158    \\
                & MonoDiffusion \cite{shao2024monodiffusion}   		&   0.297  &  \textbf{2.871}   &   6.877 &   0.156   \\
                \rowcolor{purple!10}
				& \textbf{DCPI-Depth (Ours)}		 &  	\textbf{0.291}    	&  2.944   &  \textbf{6.817}  &  \textbf{0.150}	   \\
                \hline

				\multirow{6}{*}{\makecell[c]{\textbf{DIML}}} 
                & Monodepth2 \cite{godard2019digging} & 0.185 & 0.298 & 1.140 & 0.249 \\
                & HR-Depth \cite{lyu2021hr} & 0.183 & 0.296 & 1.128 & 0.248 \\ 		
                & R-MSFM6 \cite{yan2021channel} & 0.181 & 0.301 & 1.132 & 0.243 \\
                & Lite-Mono \cite{zhang2023lite} & 0.173 & 0.271 & 1.108 & 0.239 \\
                & MonoDiffusion \cite{shao2024monodiffusion} & 0.166 & 0.256 & 1.084 & 0.232 \\

                \rowcolor{purple!10}
                & \textbf{DCPI-Depth (Ours)} & \textbf{0.163} & \textbf{0.237} & \textbf{1.038} & \textbf{0.226} \\

				\bottomrule
			\end{tabular}
		\end{center}
	\end{table}

\subsection{Generalizability Evaluation}
\label{section:general}

We incorporate the CPG stream and the BSCA strategy into existing open-source methods to further demonstrate the adaptability of our contributions to other SoTA methods. As shown in Table \ref{table:Comparisons_of_existing_models}, our proposed stream and strategy significantly improve the performance of the original networks across all metrics, which consistently validates the scalability of our methods, as detailed in Sect. \ref{section:abl}.


Finally, we conduct zero-shot experiments on the Make3D \cite{saxena2008make3d} and DIML \cite{cho2021diml} datasets using the models pretrained on the KITTI dataset. As shown in Table \ref{table:make3d}, our model outperforms all other methods across these two datasets, demonstrating its ability to generalize to new, unseen scenes.


\section{Conclusion}
\label{Sect.conclusion}
This article presented DCPI-Depth, a novel unsupervised monocular depth estimation framework with two bidirectional and collaborative streams: a conventional PCG stream and a newly developed CPG stream. The latter was designed specifically to infuse dense correspondence priors into monocular depth estimation. It consists of a CGDC loss to provide contextual-based depth with geometric guidance obtained from ego-motion and dense correspondence priors, and a DPC loss to constrain the local depth variation using the explicit relationship between the differential properties of depth and optical flow. Moreover, a BSCA strategy was developed to enhance the interaction between the two flow types, encouraging the rigid flow towards more accurate correspondence and making the optical flow more adaptable across various scenarios under the static scene hypotheses. Compared to previous works, our DCPI-Depth framework has demonstrated impressive performance and superior generalizability across six public datasets. 
Future work will focus on exploring more advanced optical flow estimation techniques to enhance the reliability and generalizability of dense correspondence priors and to extend the techniques in DCPI-Depth into a joint learning framework where multiple tasks can explicitly couple and mutually enhance each other.



\normalem
\bibliographystyle{IEEEtran}
\bibliography{ref}

\begin{thebibliography}{10}
\providecommand{\url}[1]{#1}
\csname url@samestyle\endcsname
\providecommand{\newblock}{\relax}
\providecommand{\bibinfo}[2]{#2}
\providecommand{\BIBentrySTDinterwordspacing}{\spaceskip=0pt\relax}
\providecommand{\BIBentryALTinterwordstretchfactor}{4}
\providecommand{\BIBentryALTinterwordspacing}{\spaceskip=\fontdimen2\font plus
\BIBentryALTinterwordstretchfactor\fontdimen3\font minus
  \fontdimen4\font\relax}
\providecommand{\BIBforeignlanguage}[2]{{%
\expandafter\ifx\csname l@#1\endcsname\relax
\typeout{** WARNING: IEEEtran.bst: No hyphenation pattern has been}%
\typeout{** loaded for the language `#1'. Using the pattern for}%
\typeout{** the default language instead.}%
\else
\language=\csname l@#1\endcsname
\fi
#2}}
\providecommand{\BIBdecl}{\relax}
\BIBdecl

\bibitem{geiger2013vision}
A.~Geiger \emph{et~al.}, ``{Vision meets robotics: The KITTI dataset},''
  \emph{The International Journal of Robotics Research}, vol.~32, no.~11, pp.
  1231--1237, 2013.

\bibitem{luo2020consistent}
X.~Luo \emph{et~al.}, ``Consistent video depth estimation,'' \emph{ACM
  Transactions on Graphics}, vol.~39, no.~4, pp. 71--1, 2020.

\bibitem{wang2024visual}
W.~Wang \emph{et~al.}, ``Visual robotic manipulation with depth-aware
  pretraining,'' \emph{arXiv preprint arXiv:2401.09038}, 2024.

\bibitem{zhou2017unsupervised}
T.~Zhou \emph{et~al.}, ``Unsupervised learning of depth and ego-motion from
  video,'' in \emph{Proceedings of the IEEE/CVF Conference on Computer Vision
  and Pattern Recognition (CVPR)}, 2017, pp. 1851--1858.

\bibitem{eigen2014depth}
D.~Eigen \emph{et~al.}, ``Depth map prediction from a single image using a
  multi-scale deep network,'' \emph{Advances in Neural Information Processing
  Systems (NeurIPS)}, vol.~27, 2014.

\bibitem{liu2015learning}
F.~Liu \emph{et~al.}, ``Learning depth from single monocular images using deep
  convolutional neural fields,'' \emph{IEEE Transactions on Pattern Analysis
  and Machine Intelligence}, vol.~38, no.~10, pp. 2024--2039, 2015.

\bibitem{zhou2021self}
H.~Zhou \emph{et~al.}, ``Self-supervised monocular depth estimation with
  internal feature fusion,'' \emph{arXiv preprint arXiv:2110.09482}, 2021.

\bibitem{zhao2022monovit}
C.~Zhao \emph{et~al.}, ``{MonoViT: Self-supervised monocular depth estimation
  with a vision transformer},'' in \emph{2022 International Conference on 3D
  Vision (3DV)}.\hskip 1em plus 0.5em minus 0.4em\relax IEEE, 2022, pp.
  668--678.

\bibitem{godard2019digging}
C.~Godard \emph{et~al.}, ``Digging into self-supervised monocular depth
  estimation,'' in \emph{Proceedings of the IEEE/CVF International Conference
  on Computer Vision (ICCV)}, 2019, pp. 3828--3838.

\bibitem{xu2021multi}
X.~Xu \emph{et~al.}, ``Multi-scale spatial attention-guided monocular depth
  estimation with semantic enhancement,'' \emph{IEEE Transactions on Image
  Processing}, vol.~30, pp. 8811--8822, 2021.

\bibitem{sun2023sc}
L.~Sun \emph{et~al.}, ``{SC-DepthV3: Robust self-supervised monocular depth
  estimation for dynamic scenes},'' \emph{IEEE Transactions on Pattern Analysis
  and Machine Intelligence}, vol.~46, no.~1, pp. 497--508, 2023.

\bibitem{zhang2023lite}
N.~Zhang \emph{et~al.}, ``{Lite-Mono: A lightweight CNN and Transformer
  architecture for self-supervised monocular depth estimation},'' in
  \emph{Proceedings of the IEEE/CVF Conference on Computer Vision and Pattern
  Recognition (CVPR)}, 2023, pp. 18\,537--18\,546.

\bibitem{godard2017unsupervised}
C.~Godard \emph{et~al.}, ``Unsupervised monocular depth estimation with
  left-right consistency,'' in \emph{Proceedings of the IEEE/CVF Conference on
  Computer Vision and Pattern Recognition (CVPR)}, 2017, pp. 270--279.

\bibitem{ye2021unsupervised}
X.~Ye \emph{et~al.}, ``Unsupervised monocular depth estimation via recursive
  stereo distillation,'' \emph{IEEE Transactions on Image Processing}, vol.~30,
  pp. 4492--4504, 2021.

\bibitem{teed2020raft}
Z.~Teed and J.~Deng, ``{RAFT: Recurrent all-pairs field transforms for optical
  flow},'' in \emph{Proceedings of the European Conference on Computer Vision
  (ECCV)}.\hskip 1em plus 0.5em minus 0.4em\relax Springer, 2020, pp. 402--419.

\bibitem{hartley1997triangulation}
R.~I. Hartley and P.~Sturm, ``Triangulation,'' \emph{Computer Vision and Image
  Understanding}, vol.~68, no.~2, pp. 146--157, 1997.

\bibitem{cao2017estimating}
Y.~Cao \emph{et~al.}, ``Estimating depth from monocular images as
  classification using deep fully convolutional residual networks,'' \emph{IEEE
  Transactions on Circuits and Systems for Video Technology}, vol.~28, no.~11,
  pp. 3174--3182, 2017.

\bibitem{hu2019revisiting}
J.~Hu \emph{et~al.}, ``Revisiting single image depth estimation: Toward higher
  resolution maps with accurate object boundaries,'' in \emph{2019 IEEE Winter
  Conference on Applications of Computer Vision (WACV)}.\hskip 1em plus 0.5em
  minus 0.4em\relax IEEE, 2019, pp. 1043--1051.

\bibitem{fu2018deep}
H.~Fu \emph{et~al.}, ``Deep ordinal regression network for monocular depth
  estimation,'' in \emph{Proceedings of the IEEE/CVF Conference on Computer
  Vision and Pattern Recognition (CVPR)}, 2018, pp. 2002--2011.

\bibitem{bhat2021adabins}
S.~F. Bhat \emph{et~al.}, ``{AdaBins: Depth estimation using adaptive bins},''
  in \emph{Proceedings of the IEEE/CVF Conference on Computer Vision and
  Pattern Recognition (CVPR)}, 2021, pp. 4009--4018.

\bibitem{lee2019big}
J.~H. Lee \emph{et~al.}, ``From big to small: Multi-scale local planar guidance
  for monocular depth estimation,'' \emph{arXiv preprint arXiv:1907.10326},
  2019.

\bibitem{yang2021transformer}
G.~Yang \emph{et~al.}, ``Transformer-based attention networks for continuous
  pixel-wise prediction,'' in \emph{Proceedings of the IEEE/CVF International
  Conference on Computer Vision (ICCV)}, 2021, pp. 16\,269--16\,279.

\bibitem{yang2024depth}
L.~Yang \emph{et~al.}, ``Depth anything: Unleashing the power of large-scale
  unlabeled data,'' \emph{arXiv preprint arXiv:2401.10891}, 2024.

\bibitem{birkl2023midas}
R.~Birkl \emph{et~al.}, ``Midas v3. 1--a model zoo for robust monocular
  relative depth estimation,'' \emph{arXiv preprint arXiv:2307.14460}, 2023.

\bibitem{oquab2024dinov}
\BIBentryALTinterwordspacing
M.~Oquab \emph{et~al.}, ``{DINOv2}: Learning robust visual features without
  supervision,'' \emph{Transactions on Machine Learning Research}, 2024.
  [Online]. Available: \url{https://openreview.net/forum?id=a68SUt6zFt}
\BIBentrySTDinterwordspacing

\bibitem{kopf2021robust}
J.~Kopf \emph{et~al.}, ``Robust consistent video depth estimation,'' in
  \emph{Proceedings of the IEEE/CVF Conference on Computer Vision and Pattern
  Recognition (CVPR)}, 2021, pp. 1611--1621.

\bibitem{garg2016unsupervised}
R.~Garg \emph{et~al.}, ``Unsupervised cnn for single view depth estimation:
  Geometry to the rescue,'' in \emph{Proceedings of the European Conference on
  Computer Vision (ECCV)}.\hskip 1em plus 0.5em minus 0.4em\relax Springer,
  2016, pp. 740--756.

\bibitem{zhang2020unsupervised}
Y.~Zhang \emph{et~al.}, ``Unsupervised multi-view constrained convolutional
  network for accurate depth estimation,'' \emph{IEEE Transactions on Image
  Processing}, vol.~29, pp. 7019--7031, 2020.

\bibitem{guizilini20203d}
V.~Guizilini \emph{et~al.}, ``{3D} packing for self-supervised monocular depth
  estimation,'' in \emph{Proceedings of the IEEE/CVF Conference on Computer
  Vision and Pattern Recognition (CVPR)}, 2020, pp. 2485--2494.

\bibitem{zhang2022self}
Y.~Zhang \emph{et~al.}, ``Self-supervised monocular depth estimation with
  multiscale perception,'' \emph{IEEE Transactions on Image Processing},
  vol.~31, pp. 3251--3266, 2022.

\bibitem{han2023self}
W.~Han \emph{et~al.}, ``Self-supervised monocular depth estimation by
  direction-aware cumulative convolution network,'' in \emph{Proceedings of the
  IEEE/CVF International Conference on Computer Vision (ICCV)}, 2023, pp.
  8613--8623.

\bibitem{zou2018df}
Y.~Zou \emph{et~al.}, ``Df-net: Unsupervised joint learning of depth and flow
  using cross-task consistency,'' in \emph{Proceedings of the European
  Conference on Computer Vision (ECCV)}, 2018, pp. 36--53.

\bibitem{yin2018geonet}
Z.~Yin and J.~Shi, ``{GeoNet: Unsupervised learning of dense depth, optical
  flow and camera pose},'' in \emph{Proceedings of the IEEE/CVF Conference on
  Computer Vision and Pattern Recognition (CVPR)}, 2018, pp. 1983--1992.

\bibitem{ranjan2019competitive}
A.~Ranjan \emph{et~al.}, ``Competitive collaboration: Joint unsupervised
  learning of depth, camera motion, optical flow and motion segmentation,'' in
  \emph{Proceedings of the IEEE/CVF Conference on Computer Vision and Pattern
  Recognition (CVPR)}, 2019, pp. 12\,240--12\,249.

\bibitem{chen2023self}
X.~Chen \emph{et~al.}, ``Self-supervised monocular depth estimation: Solving
  the edge-fattening problem,'' in \emph{Proceedings of the IEEE/CVF Winter
  Conference on Applications of Computer Vision (WACV)}, 2023, pp. 5776--5786.

\bibitem{jung2021fine}
H.~Jung \emph{et~al.}, ``Fine-grained semantics-aware representation
  enhancement for self-supervised monocular depth estimation,'' in
  \emph{Proceedings of the IEEE/CVF International Conference on Computer Vision
  (ICCV)}, 2021, pp. 12\,642--12\,652.

\bibitem{bian2021unsupervised}
J.-W. Bian \emph{et~al.}, ``Unsupervised scale-consistent depth learning from
  video,'' \emph{International Journal of Computer Vision}, vol. 129, no.~9,
  pp. 2548--2564, 2021.

\bibitem{wang2004image}
Z.~Wang \emph{et~al.}, ``Image quality assessment: from error visibility to
  structural similarity,'' \emph{IEEE Transactions on Image Processing},
  vol.~13, no.~4, pp. 600--612, 2004.

\bibitem{lyu2021hr}
X.~Lyu \emph{et~al.}, ``{HR-depth: High resolution self-supervised monocular
  depth estimation},'' in \emph{Proceedings of the AAAI Conference on
  Artificial Intelligence}, vol.~35, no.~3, 2021, pp. 2294--2301.

\bibitem{bian2021auto}
J.-W. Bian \emph{et~al.}, ``Auto-rectify network for unsupervised indoor depth
  estimation,'' \emph{IEEE Transactions on Pattern Analysis and Machine
  Intelligence}, vol.~44, no.~12, pp. 9802--9813, 2021.

\bibitem{bello2024self}
J.~L.~G. Bello \emph{et~al.}, ``Self-supervised monocular depth estimation with
  positional shift depth variance and adaptive disparity quantization,''
  \emph{IEEE Transactions on Image Processing}, 2024.

\bibitem{sun2023dynamo}
Y.~Sun and B.~Hariharan, ``{Dynamo-Depth}: Fixing unsupervised depth estimation
  for dynamical scenes,'' \emph{Advances in Neural Information Processing
  Systems (NeurIPS)}, 2023.

\bibitem{geiger2012we}
A.~Geiger \emph{et~al.}, ``{Are we ready for autonomous driving? the KITTI
  vision benchmark suite},'' in \emph{Proceedings of the IEEE/CVF Conference on
  Computer Vision and Pattern Recognition (CVPR)}.\hskip 1em plus 0.5em minus
  0.4em\relax IEEE, 2012, pp. 3354--3361.

\bibitem{caesar2020nuscenes}
H.~Caesar \emph{et~al.}, ``nuscenes: A multimodal dataset for autonomous
  driving,'' in \emph{Proceedings of the IEEE/CVF Conference on Computer Vision
  and Pattern Recognition (CVPR)}, 2020, pp. 11\,621--11\,631.

\bibitem{mei2022waymo}
J.~Mei \emph{et~al.}, ``Waymo open dataset: Panoramic video panoptic
  segmentation,'' in \emph{Proceedings of the European Conference on Computer
  Vision (ECCV)}.\hskip 1em plus 0.5em minus 0.4em\relax Springer, 2022, pp.
  53--72.

\bibitem{li2023sense}
G.~Li \emph{et~al.}, ``{SENSE}: Self-evolving learning for self-supervised
  monocular depth estimation,'' \emph{IEEE Transactions on Image Processing},
  2023.

\bibitem{shao2024monodiffusion}
S.~Shao \emph{et~al.}, ``{{MonoDiffusion}}: {{Self-Supervised Monocular Depth
  Estimation Using Diffusion Model}},'' \emph{IEEE Transactions on Circuits and
  Systems for Video Technology}, pp. 1--1, 2024.

\bibitem{li2021unsupervised}
H.~Li, A.~Gordon, H.~Zhao, V.~Casser, and A.~Angelova, ``Unsupervised
  {{Monocular Depth Learning}} in {{Dynamic Scenes}},'' in \emph{Proceedings of
  the 2020 {{Conference}} on {{Robot Learning}}}.\hskip 1em plus 0.5em minus
  0.4em\relax PMLR, 2021, pp. 1908--1917.

\bibitem{saxena2008make3d}
A.~Saxena \emph{et~al.}, ``{Make3D: Learning 3d scene structure from a single
  still image},'' \emph{IEEE Transactions on Pattern Analysis and Machine
  Intelligence}, vol.~31, no.~5, pp. 824--840, 2008.

\bibitem{cho2021diml}
J.~Cho \emph{et~al.}, ``Diml/cvl rgb-d dataset: 2m rgb-d images of natural
  indoor and outdoor scenes,'' \emph{arXiv preprint arXiv:2110.11590}, 2021.

\bibitem{deng2009imagenet}
J.~Deng \emph{et~al.}, ``{ImageNet}: A large-scale hierarchical image
  database,'' in \emph{Proceedings of the IEEE/CVF Conference on Computer
  Vision and Pattern Recognition (CVPR)}.\hskip 1em plus 0.5em minus
  0.4em\relax Ieee, 2009, pp. 248--255.

\bibitem{menze2015object}
M.~Menze and A.~Geiger, ``Object scene flow for autonomous vehicles,'' in
  \emph{Proceedings of the IEEE/CVF Conference on Computer Vision and Pattern
  Recognition (CVPR)}, 2015, pp. 3061--3070.

\bibitem{uhrig2017sparsity}
J.~Uhrig \emph{et~al.}, ``{Sparsity invariant CNNs},'' in \emph{2017
  International Conference on 3D Vision (3DV)}.\hskip 1em plus 0.5em minus
  0.4em\relax IEEE, 2017, pp. 11--20.

\bibitem{he2016deep}
K.~He \emph{et~al.}, ``Deep residual learning for image recognition,'' in
  \emph{Proceedings of the IEEE/CVF Conference on Computer Vision and Pattern
  Recognition (CVPR)}, 2016, pp. 770--778.

\bibitem{shim2023swindepth}
D.~Shim and H.~J. Kim, ``Swindepth: Unsupervised depth estimation using
  monocular sequences via swin transformer and densely cascaded network,'' in
  \emph{Proceedings of the IEEE International Conference on Robotics and
  Automation (ICRA)}.\hskip 1em plus 0.5em minus 0.4em\relax IEEE, 2023, pp.
  4983--4990.

\bibitem{yan2021channel}
J.~Yan \emph{et~al.}, ``Channel-wise attention-based network for
  self-supervised monocular depth estimation,'' in \emph{2021 International
  Conference on 3D vision (3DV)}.\hskip 1em plus 0.5em minus 0.4em\relax IEEE,
  2021, pp. 464--473.

\end{thebibliography}

\end{document}